\pdfoutput=1

\documentclass[11pt]{article}

\usepackage[preprint]{acl}

\usepackage{times}
\usepackage{latexsym}

\usepackage{algorithm}
\usepackage{algorithmic}

\usepackage{graphicx}
\usepackage{amsmath}
\usepackage{amssymb}
\usepackage{booktabs}
\usepackage{pifont}
\usepackage{times}
\usepackage{soul}
\usepackage{url}
\usepackage{multirow}
\usepackage[utf8]{inputenc}
\usepackage{wrapfig}
\usepackage{makecell}
\usepackage{listings}
\usepackage{framed}
\usepackage{textcomp}
\usepackage{minitoc}
\usepackage{colortbl}
\usepackage{xcolor}
\usepackage{algorithm}
\usepackage{algorithmic}
\usepackage{amssymb}
\usepackage{xcolor}
\usepackage{pifont}
\usepackage{booktabs}  
\usepackage{colortbl}  
\usepackage{multirow}  
\usepackage{array}
\usepackage{graphicx}  
\usepackage{xcolor}    
\usepackage{pifont}    
\usepackage{tikz}
\usepackage{minitoc}
\usepackage{booktabs}
\usepackage{siunitx}
\hypersetup{breaklinks=true}
\usepackage[most]{tcolorbox}
\usepackage{caption}
\usepackage[capitalize]{cleveref}
\crefname{section}{Sec.}{Secs.}
\Crefname{section}{Section}{Sections}
\Crefname{table}{Table}{Tables}
\crefname{table}{Tab.}{Tabs.}

\usepackage{newfloat}
\usepackage{listings}

\usepackage[T1]{fontenc}

\usepackage[utf8]{inputenc}

\usepackage{microtype}

\usepackage{inconsolata}

\usepackage{graphicx}

\title{%
  \makebox[0pt][l]{\raisebox{-0.3\height}{\includegraphics[height=1.6cm]{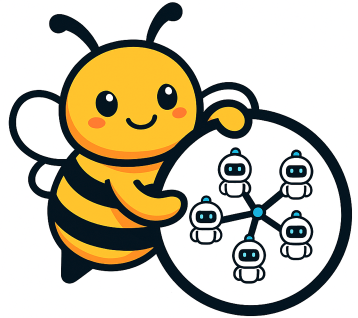}}}%
  \makebox[\dimexpr\textwidth+0.5cm\relax][c]{SwarmAgentic: Towards Fully Automated Agentic System}\\[-0.4em]
  \makebox[\dimexpr\textwidth+0.7cm\relax][c]{Generation via Swarm Intelligence}
}

\author{Yao Zhang \textsuperscript{\rm 1,3}\qquad Chenyang Lin \textsuperscript{\rm 2}
\qquad Shijie Tang \textsuperscript{\rm 1}
\qquad Haokun Chen\textsuperscript{\rm 1} \\
\qquad \bf Shijie Zhou \textsuperscript{\rm 2}
\qquad \bf Yunpu Ma \textsuperscript{\rm 1,3}
\qquad \bf Volker Tresp\textsuperscript{\rm 1,3}\\
    \textsuperscript{\rm 1} LMU Munich \qquad  
    \textsuperscript{\rm 2} Technical University of Munich \qquad 
    \textsuperscript{\rm 3} Munich Center for Machine Learning \\
    yzhang@dbs.ifi.lmu.de \qquad tresp@dbs.ifi.lmu.de
   	}

\begin{document}
\maketitle

\begin{abstract}
The rapid progress of Large Language Models has advanced agentic systems in decision-making, coordination, and task execution. Yet, existing agentic system generation frameworks lack full autonomy, missing from-scratch agent generation, self-optimizing agent functionality, and collaboration, limiting adaptability and scalability. We propose \textbf{SwarmAgentic}, a framework for \textit{fully automated agentic system generation} that constructs agentic systems from scratch and jointly optimizes agent functionality and collaboration as interdependent components through language-driven exploration. To enable efficient search over system-level structures, SwarmAgentic maintains a population of candidate systems and evolves them via feedback-guided updates, drawing inspiration from Particle Swarm Optimization (PSO). We evaluate our method on six real-world, open-ended, and exploratory tasks involving high-level planning, system-level coordination, and creative reasoning. Given only a task description and an objective function, SwarmAgentic outperforms all baselines, achieving a \textbf{+261.8\% relative improvement} over ADAS on the TravelPlanner benchmark, highlighting the effectiveness of full automation in structurally unconstrained tasks. This framework marks a significant step toward scalable and autonomous agentic system design, bridging swarm intelligence with fully automated system multi-agent generation. Our code is publicly released at \href{https://yaoz720.github.io/SwarmAgentic/}{github.com/SwarmAgentic}.
\end{abstract}

\section{Introduction}

\begin{table*}[h]
\centering
\scalebox{0.85}{
\begin{tabular}{l>{\centering\arraybackslash}m{3.3cm}
	>{\centering\arraybackslash}m{3.3cm}
	>{\centering\arraybackslash}m{3.3cm}}
\toprule
\textbf{Framework} & \textbf{\shortstack{From-Scratch \\ Agent Generation}} & \textbf{\shortstack{Self-Optimizing  \\ Agent Functionality}} & \textbf{\shortstack{Self-Optimizing \\Agent Collaboration}} \\ 
\midrule
\textbf{SPP~\cite{wang2023unleashing}} & \textcolor{red}{\ding{55}} & \textcolor{red}{\ding{55}} & \textcolor{red}{\ding{55}} \\ 
\textbf{EvoAgent~\cite{yuan2024evoagent}} & \textcolor{red}{\ding{55}} & \textcolor{green}{\checkmark} & \textcolor{red}{\ding{55}} \\ 
\textbf{AgentSquare~\cite{shang2024agentsquare}} & \textcolor{red}{\ding{55}} & \textcolor{green}{\checkmark} & \textcolor{red}{\ding{55}} \\ 
\textbf{AutoAgents~\cite{chen2023autoagents}} & \textcolor{red}{\ding{55}} & \textcolor{green}{\checkmark} &\textcolor{green}{\checkmark} \\ 
\textbf{AFlow~\cite{zhang2024aflow}} & \textcolor{red}{\ding{55}} & \textcolor{green}{\checkmark} & \textcolor{green}{\checkmark} \\ 
\textbf{Agent Sym. Learning~\cite{zhou2024symbolic}} & \textcolor{red}{\ding{55}} & \textcolor{green}{\checkmark} & \textcolor{green}{\checkmark} \\ 
\textbf{ADAS~\cite{hu2024automated}} & \textcolor{red}{\ding{55}} & \textcolor{green}{\checkmark} & \textcolor{green}{\checkmark} \\
\hline
\textbf{SwarmAgentic} & \textcolor{green}{\checkmark} & \textcolor{green}{\checkmark} & \textcolor{green}{\checkmark} \\ 
\bottomrule
\end{tabular}
}
\caption{Comparison between SwarmAgentic and existing frameworks along three dimensions of agentic system autonomy. SwarmAgentic is the only framework satisfying all three, enabling fully automated and scalable agentic system generation without human intervention. See Appendix \ref{app: Agentic Autonomy Evaluation Framework} for definitions and capability assessments.}
\label{table:benchmark-comparison}
\vspace{-5pt}
\end{table*}

The advancement of Large Language Models (LLMs)~\cite{achiam2023gpt, guo2025deepseek} has substantially advanced the capabilities of agentic systems~\cite{du2023improving, shinn2024reflexion, wang2024mixture}, enabling autonomous decision-making~\cite{li2025embodied}, coordination~\cite{qian2024scaling}, and complex task execution~\cite{xi2024agentgym, Zhang_Ma_Ma_Han_Wu_Tresp_2025}. Nonetheless, current agentic system generation frameworks lack full autonomy, missing from-scratch agent generation, self-optimizing functionality, and collaboration~\cite{wu2023autogen, li2023camel, hong2023metagpt}. These design rigidities limit adaptability and scalability, suppress the emergence of self-optimizing system behaviors, and impose significant engineering overhead. As a result, such systems struggle to accommodate diverse and complex task specifications without substantial manual intervention.

This challenge becomes even more pronounced in open-ended, exploratory tasks that require high-level planning and system-level coordination, where manually designing agents and their collaboration strategies is prohibitively complex, labor-intensive, and hard to scale. Addressing this requires a practical framework equipped with three core capabilities: \textit{From-Scratch Agent Generation, Self-Optimizing Agent Functionality, and Self-Optimizing Agent Collaboration}, together enabling scalable and fully autonomous agentic system construction. While recent work has explored agentic system automation~\cite{khattab2023dspy, zhangoffline, wang2023unleashing}, no existing framework satisfies all three autonomy criteria, as shown in Tab.~\ref{table:benchmark-comparison}. SPP~\cite{wang2023unleashing} lacks from-scratch agent generation, behavior adaptation, and collaboration restructuring. EvoAgent~\cite{yuan2024evoagent} and AgentSquare~\cite{shang2024agentsquare} support functionality optimization but rely on fixed structures. AutoAgents~\cite{chen2023autoagents}, AFlow~\cite{zhang2024aflow}, Agent Symbolic Learning~\cite{zhou2024symbolic}, and ADAS~\cite{hu2024automated} depend on templates or seed agents, and thus fail to generate agents from scratch.

To address this gap, we introduce \textbf{SwarmAgentic}, a fully automated framework for agentic system generation that explores a language-driven, symbolic design space to instantiate agents from scratch and jointly optimize their functionalities and collaboration strategies, entirely without human intervention. To support this search process, SwarmAgentic leverages a gradient-free, population-based optimization scheme inspired by Particle Swarm Optimization (PSO)~\cite{Kennedy2002Particle}, which is well-suited for navigating non-differentiable, structurally diverse system configurations through parallel exploration and iterative refinement.

Specifically, SwarmAgentic represents each agentic system as a particle, encoding agents and their collaboration strategies in structured language. Unlike traditional PSO, which optimizes numerical vectors, SwarmAgentic employs language-based transformations for velocity and position updates, ensuring interpretable optimization. The process begins with particle initialization to generate diverse agentic systems, followed by LLM-driven flaw identification to detect inefficiencies. Velocity updates integrate failure-driven adjustments, personal best guidance, and global best guidance to balance self-learning and swarm-based improvements. Position updates iteratively refine system configurations until a stopping criterion is met.

We evaluate SwarmAgentic on six real-world, open-ended, and exploratory tasks that demand high-level planning, system-level coordination, and creative reasoning. Given only a task description and an objective function, SwarmAgentic achieves a \textbf{261.8\% relative gain} over ADAS on TravelPlanner, and outperforms all baselines across Trip Planning, Meeting Planning, Calendar Scheduling, Creative Writing, and MGSM. These results highlight the effectiveness of fully automated agentic system generation on structurally unconstrained tasks, where no fixed templates or handcrafted agents can be reused. This framework marks a significant step toward scalable and autonomous agentic system design, bridging swarm intelligence with fully automated system generation.

The key contributions of this work are:\\
1. We introduce SwarmAgentic, a fully automated framework for agentic system generation that requires no predefined agents or human intervention. It leverages a language-driven population-based search to jointly optimize agent functionality and collaboration.\\
2.	We reformulate PSO into a symbolic, language-based optimization process, where agents and their coordination strategies are encoded as structured representations and evolved in a non-differentiable design space.\\
3.	We propose a Failure-Aware Velocity Update mechanism, which incorporates LLM-guided flaw identification to dynamically guide configuration refinement, enabling targeted self-optimization across iterations.\\
4.	We demonstrate that SwarmAgentic outperforms strong baselines on six real-world, open-ended tasks involving high-level planning and multi-agent coordination, achieving state-of-the-art performance with only task descriptions and objective functions as input.

\section{Related Work}

\subsection{Agentic System Generation}
LLM-based multi-agent frameworks~\cite{li2023camel, wu2023autogen, hong2023metagpt} enhance task-solving through agent collaboration but rely on fixed workflows and human intervention, limiting adaptability. Recent approaches, such as SPP~\cite{wang2023unleashing} and AgentVerse~\cite{chen2023agentverse}, automate large-scale agent generation—SPP simulates multi-persona collaboration, while AgentVerse assembles expert teams. AutoAgents~\cite{chen2023autoagents} refines agents through discussion-driven iteration, and EvoAgent~\cite{yuan2024evoagent} optimizes multi-agent configurations via evolutionary algorithms. Despite progress in automation, these methods treat agent collaboration strategies as static templates, restricting adaptability. In contrast, SwarmAgentic eliminates predefined constraints by jointly optimizing agent functionality and collaboration strategies through a language-driven PSO framework, enabling fully automated and scalable agentic system generation.

\subsection{Agentic System Optimization}
Optimizing agentic systems requires refining both agent functionalities and collaboration strategies. In single-agent settings, methods like Agent-Pro~\cite{zhang2024agentprolearningevolvepolicylevel} improve agent policies through trajectory-based updates, while multi-agent approaches, such as GPTSwarm~\cite{zhuge2024language} and DyLAN~\cite{liu2024dynamic}, focus on optimizing inter-agent coordination. AUTOACT~\cite{qiao2024autoact} refines agent decisions through filtered trajectories, while AutoFlow~\cite{li2024autoflow} leverages reinforcement learning for workflow optimization. ADAS~\cite{hu2024automated} and AgentSquare~\cite{shang2024agentsquare} further enhance adaptability by exploring diverse system module compositions. Additionally, Agent Symbolic Learning~\cite{zhou2024symbolic} and G\"odelAgent~\cite{yin2024g} leverage text-based gradient optimization for recursive self-improvement. However, these methods separately optimize agent functionality and collaboration, limiting adaptability. SwarmAgentic unifies both as interdependent components, using language-driven PSO to dynamically refine agentic systems.

\begin{figure*}[th]
    \centering
    \includegraphics[width=1\linewidth]{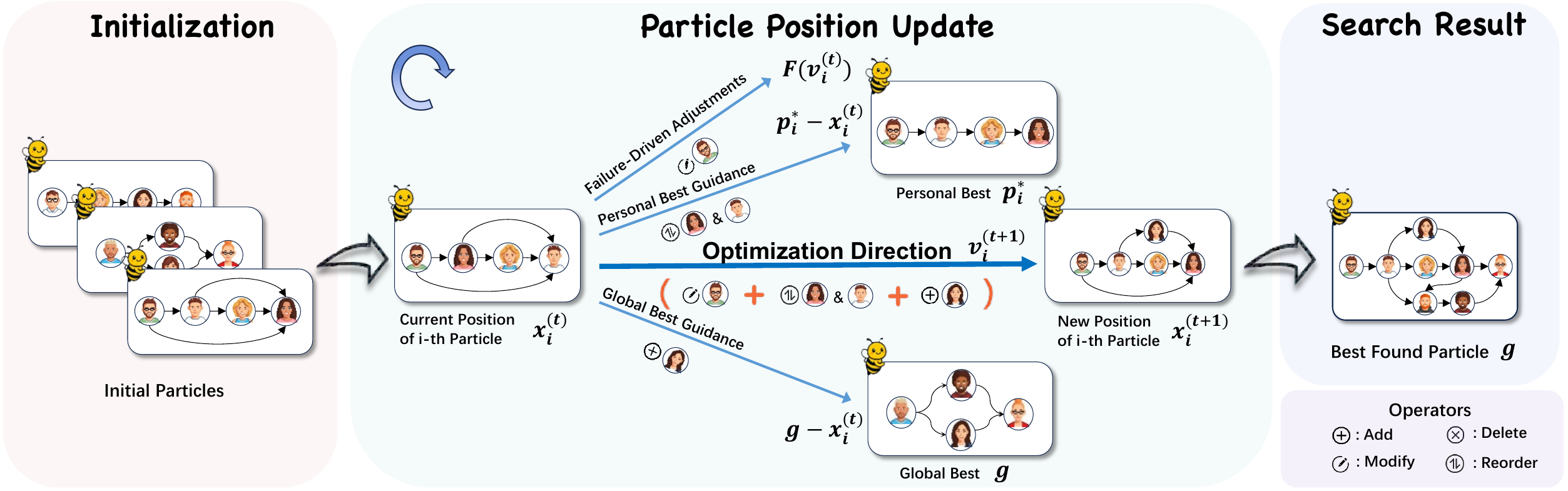}
    \caption{Overview of SwarmAgentic. (1) Initialization: Generates a diverse population of agentic systems, encoding agent sets, and collaboration structures in a structured language space. (2) Particle Position Update: Iteratively refines agentic systems through failure-aware velocity updates and position updates, incorporating failure-driven adjustments, personal best guidance, and global best guidance. Both velocity and position updates operate on structured language representations, enabling interpretable transformations over agent functionality and collaboration strategies (see Appendix \ref{app: Optimization Mechanism Illustration} for examples). (3) Search Result: Returns the best-performing agentic system $g$, refined through structured updates that balance self-adaptation and swarm-based optimization for enhanced coordination and efficiency. The pseudo code for SwarmAgentic is in Appendix \ref{app: Pseudo Code for SwarmAgentic}}
    \label{fig_main}
    \vspace{-5pt}
\end{figure*}

\section{Preliminary}

\subsection{Agentic System Optimization}
An agentic system at generation \( t \), denoted as \( \mathcal{S}_i^{(t)} \), represents the \( i \)-th solution within the population. It comprises an agent set \( \mathcal{A}_i^{(t)} = \{A_{i,1}^{(t)}, A_{i,2}^{(t)}, \dots, A_{i,m}^{(t)} \} \) and a collaborative structure \( \mathcal{W}_i^{(t)} \). Each agent \( A_{i,k}^{(t)} \), where \( k \in \{1, \dots, m\} \), is represented as: $
A_{i,k}^{(t)} = \left( I_{i,k}^{(t)}, R_{i,k}^{(t)}, P_{i,k}^{(t)} \right),
$ where \( I_{i,k}^{(t)} \) is the agent identifier, uniquely defining its role within the system, \( R_{i,k}^{(t)} \) is the responsibility, specifying the tasks it is capable of performing, and \( P_{i,k}^{(t)} \) is the execution policy, governing its decision-making and task execution. Agents operate within a collaborative structure, defined as:
$ \mathcal{W}_i^{(t)} = \{ W_{i, 1}^{(t)}, W_{i, 2}^{(t)}, \dots, W_{i, n}^{(t)} \},$ where each step \( W_{i, l}^{(t)} \), \( l \in \{1, \dots, n\} \) assigns a specific agent \( A_{i,k}^{(t)} \) to execute the corresponding task. SwarmAgentic iteratively refines the agent set \( \mathcal{A} \) and collaborative structures \( \mathcal{W} \) to optimize the agentic system \( \mathcal{S} \), aiming to maximize task performance, which is quantitatively assessed by the fitness function \( J(\mathcal{S}) \). The Basic Structure of the Agentic System is detailed in Appendix \ref{app: Basic Structure of Agentic System}.

\subsection{Particle Swarm Optimization}
PSO~\cite{Kennedy2002Particle}, inspired by swarm intelligence, models the dynamic adaptation processes observed in natural systems, such as bird flocking and fish schooling. Each particle iteratively refines its position based on individual experiences while incorporating shared knowledge from the swarm, balancing exploration and exploitation. This decentralized and self-organizing mechanism makes PSO particularly well-suited for optimization in complex search spaces. Each particle maintains a position \( x_i^{(t)} \), representing a candidate solution, and a velocity \( v_i^{(t)} \), which updates its movement in the search space. The position and velocity updates follow:
\begin{equation}
\resizebox{0.89\linewidth}{!}{$
v_i^{(t+1)} = \omega v_i^{(t)} + c_1 r_1 (p_i^* - x_i^{(t)}) + c_2 r_2 (g - x_i^{(t)}),
$}
\end{equation}
\begin{equation}
x_i^{(t+1)} = x_i^{(t)} + v_i^{(t+1)},
\end{equation}
where \( p_i^* \) is the personal best found by particle \( i \), and \( g \) is the global best in the swarm. The inertia weight \( \omega \) balances exploration and exploitation, while the learning coefficients \( c_1, c_2 \) determine the influence of personal and global bests. The stochastic factors \( r_1, r_2 \) introduce randomness to enhance diversity and prevent premature convergence. After each iteration, each particle is evaluated using the fitness function \( J \), guiding the optimization process until a predefined stopping criterion, such as a fixed number of iterations, is met. Unlike traditional PSO on continuous vectors, our setting optimizes discrete, structured configurations. We reinterpret velocity and position updates as semantic transformations over language-based representations, enabling swarm-based search in symbolic, high-dimensional language spaces.

\section{SwarmAgentic}
\label{sec: SwarmAgentic}
SwarmAgentic adapts PSO to a language-based search space, optimizing agentic systems as structured textual representations. Unlike traditional PSO in continuous vector spaces, it explores a combinatorial space of agent functionalities and collaboration strategies. Each particle encodes an agentic system in language, and position updates are realized as text-based transformations guided by structural feedback, enabling population-based search in discrete, non-numeric domains.

The optimization process begins with \textit{particle initialization}, where candidate agentic systems are randomly synthesized from the task description using an LLM. Unlike numerical optimization, where position updates are directly guided by fitness scores, SwarmAgentic first performs \textit{flaw identification} by analyzing system performance against the objective function, identifying inefficiencies before making adjustments to ensure targeted optimization. Building on flaw identification, SwarmAgentic generates optimization directions through \textit{failure-aware velocity updates}, integrating failure-driven adjustments, personal best guidance (self-learning), and global best guidance (swarm-based). The refinements from velocity updates are applied through \textit{position updates}, modifying agent functionalities and collaboration strategies. By translating optimization directions into concrete adjustments, position updates iteratively refine agentic system configurations until the predefined iteration limit is met. The best-performing system $g$ is retained as the final solution. The following sections detail each step, illustrating how SwarmAgentic transitions from numerical-based updates to language-driven transformations for structured and interpretable optimization.

\subsection{Particle Initialization}
\label{sec:init}
SwarmAgentic initializes a diverse population of candidate agentic system \( \mathcal{S}_i^{(0)} \) each represented as a particle in the PSO search space. A system comprises a collaborative structure \( \mathcal{W}_i^{(0)} \) and an agent set \( \mathcal{A}_i^{(0)} \), with its configuration encoded as an initial position \( x_i^{(0)} \). The velocity \( v_i^{(0)} \) governs iterative textual modifications, progressively refining \( \mathcal{A}_i \) and \( \mathcal{W}_i \) throughout the optimization process.

To enhance structural diversity, we employ a temperature-controlled sampling strategy. Specifically, low-temperature particles generate stable configurations closely aligned with established patterns. Medium-temperature particles introduce moderate variability, balancing structural stability and design innovation. High-temperature particles maximize exploration, yielding unconventional architectures that expand the search space. This stratification balances exploitation of high-performing structures with exploration of novel solutions.

Velocity initialization influences early search by directing particles toward promising regions while maintaining diversity. Initial velocities are assigned based on estimated fitness, promoting convergence while preventing stagnation in suboptimal configurations. The personal best of each particle is set to its initial position, with fitness evaluated using predefined task-specific metrics. The global best remains undefined until all particles are assessed, after which the top-performing configuration serves as a reference for subsequent optimization.

\subsection{Flaw Identification}
\label{sec:flaw-identification}
In language-driven optimization frameworks, identifying flaws is essential to ensure refinements are targeted and effective. Unlike traditional PSO, which updates positions based on scalar fitness scores, SwarmAgentic detects system deficiencies through an LLM-driven analysis of execution failures, enabling structured and interpretable updates. Flaws in agentic systems can be categorized into agent flaws and collaborative structures flaws, both of which impact efficiency and reliability. Agent flaws include missing agents that leave critical tasks unassigned, redundant agents that introduce inefficiencies, and ambiguous policies that hinder coordination. Collaborative structures flaws encompass missing steps that disrupt execution, redundant steps that increase overhead, incomplete contextual information that prevents agents from making informed decisions, and misaligned task outcomes that propagate errors to subsequent steps, leading to cascading failures. SwarmAgentic systematically identifies system deficiencies through structured evaluation. Task performance is assessed based on the objective function, producing an error set \( \mathcal{E}_i^{(t)} \). Given \( \mathcal{E}_i^{(t)} \) and the current system \( \mathcal{S}_i^{(t)} \), an LLM analyzes failure patterns and derives flaw \( f_i^{(t+1)} \), which consists of agent flaws and structures flaws. This structured diagnosis ensures that velocity updates are informed by actual performance bottlenecks rather than arbitrary modifications, leading to more effective system refinements.

\begin{table*}[t]
\centering
\scalebox{0.88}{
\begin{tabular}{lcccccc}
    \toprule
	 \multirow{2}{*}{\textbf{Method}} & \textbf{Delivery} & \multicolumn{2}{c}{\textbf{Commonsense}} & \multicolumn{2}{c}{\textbf{Hard Constraint}} & \multirow{2}{*}{\textbf{Final}} \\
    \cmidrule(lr){3-4} \cmidrule(lr){5-6}
    & \textbf{Rate} & \textbf{Micro} & \textbf{Macro} & \textbf{Micro} & \textbf{Macro} &\\ 
    \midrule
    Direct & 100.0 / 100.0 & 57.3 / 79.4 & 3.9 / 15.8 & 11.0 / 27.5 & 3.3 / 16.1 & 0.0 / 2.2 \\
    CoT~\cite{wei2022chain} & 100.0 / 100.0 & 61.0 / 76.7 & 2.8 / 11.7 & 10.0 / 22.4 & 3.3 / 12.8 & 0.0 / 2.2 \\
    Self-Refine~\cite{madaan2024self} & 100.0 / 98.9 & 56.0 / 75.3 & 1.7 / 7.2 & 3.1 / 12.4 & 1.1 / 7.2 & 0.0 / 1.1 \\
    SPP~\cite{wang2023unleashing} & 99.4 / 96.7 & 54.6 / 70.6 & 1.7 / 5.6 & 3.8 / 11.4 & 1.1 / 7.8 & 0.0 / 0.6 \\
    EvoAgent~\cite{yuan2024evoagent} & 100.0 / 100.0 & 64.2 / 81.5 & 7.8 / 21.1 & 11.0 / 31.4 & 4.4 / 18.9 & 1.1 / 7.2 \\
    ADAS~\cite{hu2024automated} & 100.0 / 100.0 & 70.9 / 88.5 & 6.1 / 34.4 & 17.4 / 50.2 & 9.4 / 27.8 & 1.1 / 8.9 \\
    \midrule
    \textbf{SwarmAgentic} & 100.0 / 100.0 & \textbf{70.9} / \textbf{92.9} & \textbf{12.8} / \textbf{56.1} & \textbf{21.0} / \textbf{66.7} & \textbf{9.4} / \textbf{52.8} & \textbf{3.3} / \textbf{32.2} \\
    \bottomrule
\end{tabular}}
\caption{Performance on the TravelPlanner. Each cell shows results in the format: GPT-3.5 / GPT-4o. SwarmAgentic outperforms all baseline methods, highlighting its effectiveness in automated agentic system generation.}
\label{tab:tp-result}
\vspace{-5pt}
\end{table*}

\subsection{Failure-Aware Velocity Update}

SwarmAgentic enhances traditional PSO by incorporating memory-based adaptation and language-driven velocity updates, structuring refinements as textual transformations rather than numerical adjustments. SwarmAgentic leverages an LLM to perform failure-aware refinements, enabling precise corrections rather than indiscriminately reinforcing past configurations. By integrating \textit{failure-driven adjustments, personal best guidance, and global best guidance}, SwarmAgentic systematically eliminates recurring flaws, ensuring that velocity updates lead to meaningful structural improvements. The velocity update follows:
\begin{equation}
\begin{split}
v_i^{(t+1)} & =\text{LLM}_{\text{vel}}
( c_f r_f  F(v_i^{(t)}), \\
&\quad c_p r_p (p_i^* - x_i^{(t)}), c_g r_g (g - x_i^{(t)})),
\end{split}
\label{eq:update_vel}
\end{equation}
where $c_f, c_p, c_g$ represent the repulsion coefficient, cognitive coefficient, and social coefficient, respectively, governing \textit{failure-driven adjustments,  personal best guidance, and global best guidance}. $r_f, r_p, r_g$ are stochastic exploration factors, introducing controlled randomness to enhance search diversity. $F(v_i^{(t)})$ encapsulates failure-driven adjustments, identifying the failed component of the previous velocity update.

\paragraph{Failure-Driven Adjustments.}
SwarmAgentic records failed modifications and uses LLM-based refinement to eliminate ineffective updates. The failure experience term captures unsuccessful velocity updates that did not improve task performance. Integrated into the velocity update, this memory mechanism prevents repeated suboptimal adjustments. To refine updates, SwarmAgentic provides the LLM with identified flaws from the previous configuration $f_i^{(t)}$, current configuration $f_i^{(t+1)}$, and prior update plan $v_i^{(t)}$. By analyzing these inputs, the LLM detects persistent flaws and ineffective corrections, refining velocity updates as:
\begin{equation}
c_f r_f  F(v_i^{(t)}) = \text{LLM}_{\text{fail}}
(v_i^{(t)}, f_i^{(t)}, f_i^{(t+1)}).
\label{eq:faliure}
\end{equation}

\paragraph{Personal Best Guidance.}  
Each particle retains its highest-performing configuration as a personal best $p_i^*$. Instead of directly following $p_i^*$, SwarmAgentic utilizes an LLM to compare the current configuration $x_i^{(t)}$ with $p_i^*$, refining updates based on the identified flaws $f_i^{(t+1)}$ to ensure precise corrections. Formally,
\begin{equation}
c_p  r_p (p_i^* - x_i^{(t)}) = \text{LLM}_{\text{pers}}
(x_i^{(t)}, p_i^*, f_i^{(t+1)}).
\label{eq:personal_best}
\end{equation}  

\paragraph{Global Best Guidance.}  

Each particle references the highest-performing configuration in the swarm as the global best $g$, guiding updates while balancing exploration and exploitation to prevent premature convergence. Instead of directly following $g$, SwarmAgentic employs an LLM to refine updates by comparing the current configuration $x_i^{(t)}$ with $g$ and identifying transferable improvements based on detected flaws $f_i^{(t+1)}$. Formally,
\begin{equation}
c_g  r_g (g - x_i^{(t)}) = \text{LLM}_{\text{glob}}
(x_i^{(t)}, g, f_i^{(t+1)}).
\label{eq:global_best}
\end{equation}

\subsection{Position Update}
After updating velocity, each agentic system applies structural transformations to refine its configuration as follows:
\begin{equation}
x_i^{(t+1)} = \text{LLM}_{\text{pos}}
(x_i^{(t)}, v_i^{(t+1)}).
\label{eq:position_update}
\end{equation}

SwarmAgentic optimizes agentic systems through two key adaptation mechanisms: (1) Agent-Level Adaptation: Modifies individual agents \( A_{i,k}^{(t)}\) by adjusting roles \( I_{i,k}^{(t)}\), responsibility \( R_{i,k}^{(t)}\), and execution policies \( P_{i,k}^{(t)}\) to enhance performance. New agents may be introduced, while redundant ones are removed based on feedback. (2) Collaborative Structures Reconfiguration: Enhances the collaborative structures \( \mathcal{W}_i^{(t)} \) by optimizing task sequencing, refining dependencies, and improving inter-agent coordination. Steps are reordered to streamline execution, redundant ones eliminated to reduce overhead, and new steps incorporated as necessary. Through iterative refinement, SwarmAgentic continuously improves agent functionality and collaborative structures, ensuring efficiency, adaptability, and structural coherence across generations.

\begin{table*}[ht]
\centering
\scalebox{0.86}{
\begin{tabular}{lccccc}
    \toprule
    \multirow{2}{*}{\textbf{Method}} & \multicolumn{3}{c}{\textbf{Natural Plan (NP)}} & \multirow{2}{*}{\textbf{Creative Writing (CW)}} & \multirow{2}{*}{\textbf{MGSM}} \\
    \cmidrule(lr){2-4}
    & \textbf{\shortstack{Trip \\Planning}}& \textbf{\shortstack{Meeting \\Planning}}& 
    \textbf{\shortstack{Calendar \\Scheduling}}&  &  \\ 
    \midrule
    Direct       & 7.3 / 3.7  & 19.0 / 45.0 & 19.9 / 43.0 & 5.0 / 6.3 & 28.1 / 87.3 \\
    CoT~\cite{wei2022chain}   & 9.0 / 1.0  & 19.0 / 50.0 & 20.0 / 60.0 & 5.3 / 7.0 & 28.7 / 81.0 \\
    Self-Refine~\cite{madaan2024self} & 4.4 / 4.4 & 12.0 / 41.0 & 13.0 / 63.0 & 5.2 / 6.2 & 30.5 / 86.4 \\
    SPP~\cite{wang2023unleashing}      & 5.0 / 1.3   & 4.0 / 33.0 & 22.0 / 44.0 & 5.9 / 7.6 & 55.2 / 84.9 \\
    EvoAgent~\cite{yuan2024evoagent}   & 5.6 / 1.9 & 4.0 / 38.0 & 21.6 / 52.0 & 6.1 / 7.1 & 57.3 / 87.0 \\
    ADAS~\cite{hu2024automated}        & 1.9 / 3.1 & 11.0 / 43.0 & 21.0 / 66.0 & 6.2 / 7.3 & 29.0 / 87.0 \\
    \midrule
    \textbf{SwarmAgentic} & \textbf{13.1} / \textbf{13.1} & \textbf{23.0} / \textbf{56.0} & \textbf{28.0} / \textbf{82.0} & \textbf{8.2} / \textbf{8.5} & \textbf{65.6} / \textbf{88.4} \\
    \bottomrule
\end{tabular}}
\caption{Performance on Natural Plan, Creative Writing, and MGSM. Results are shown as GPT-3.5 / GPT-4o. SwarmAgentic achieves the highest performance across all tasks, significantly outperforming baseline methods.}
\label{tab:natural_plan_creative}
\vspace{-5pt}
\end{table*}

\section{Experiments}
\subsection{Experimental Setup}

\paragraph{Tasks}
We evaluate SwarmAgentic on six real-world tasks spanning planning, collaboration, generation, and reasoning. Most are open-ended and structurally unconstrained, requiring high-level planning, system-level coordination. Specifically, we consider: (1) \textit{TravelPlanner (TP)}~\cite{xie2024travelplanner}, which tests long-horizon planning under user-defined constraints; (2–4) \textit{Trip Planning}, \textit{Meeting Planning}, and \textit{Calendar Scheduling} from \textit{Natural Plan (NP)}~\cite{zheng2024natural}, which involve multi-agent scheduling with conflict minimization; (5) \textit{Creative Writing (CW)}~\cite{yao2024tree}, which requires coherent multi-paragraph generation from unordered key points. These tasks challenge predefined agent templates due to their structural variability and open-ended semantics. Additionally, we include (6) \textit{MGSM}~\cite{shi2022language}, a structured math reasoning task where predefined logic may suffice, to evaluate generalization to template-compatible domains. Dataset details and evaluation metrics are in Appendix \ref{app: Dataset Statistics and Evaluation}

\paragraph{Baselines}
We compare SwarmAgentic with both standard prompting methods and automated approaches for agentic system generation. The prompting baselines include: (1) Direct, where the model responds with a fixed prompt; (2) CoT~\cite{wei2022chain}, which improves reasoning via step-by-step generation; and (3) Self-Refine~\cite{madaan2024self}, which iteratively refines outputs through self-feedback. For automated agentic systems, we select methods that minimize task-specific priors to reduce human intervention and better expose the underlying capacity for autonomous agent discovery, including: (4) SPP~\cite{wang2023unleashing}, which performs multi-turn self-collaboration across multiple personas; (5) EvoAgent~\cite{yuan2024evoagent}, which evolves agent configurations via optimization over roles, prompts, and behavior policies; and (6)ADAS~\cite{hu2024automated}, which uses a meta agent to discover agentic systems in code through iterative generation and refinement. Detailed baseline implementations are in Appendix \ref{app: Baseline Implementations and Configurations}.

\paragraph{Models and Implementation Details} 
SwarmAgentic, following ADAS, employs distinct models for optimization and execution. Specifically, we use GPT-4o-mini-0718~\cite{openai2024gpt4omini} as the optimizer, and select GPT-3.5-turbo-0125~\cite{openai2022chatgpt}, GPT-4o-0806~\cite{openai_gpt4o}, Claude-3.5-sonnet-0620~\cite{claude35sonnet2024}, DeepSeek-V3~\cite{deepseekai2024deepseekv3}, Gemini-1.5-Pro~\cite{pichai2024gemini15} as executor models. SwarmAgentic is configured with 5 particles and 10 optimization iterations, while ADAS is run with a maximum of 30 iterations.

\begin{table*}[ht]
\centering
\scalebox{0.88}{
\begin{tabular}{lcccccc}
\toprule
\textbf{Method} & \textbf{GPT-4o} & \textbf{Claude-3.5-sonnet} & \textbf{DeepSeek-V3} & \textbf{Gemini-1.5} & \textbf{Gemini-1.5\textbf{*}} \\
\midrule
Direct & 6.3 & 5.6 & 6.4 & 5.4 & - \\
CoT~\cite{wei2022chain} & 7.0 & 5.7 & 5.9 & 5.8 & - \\
Self-Refine~\cite{madaan2024self} & 6.2 & 5.8 & 6.1 & 5.4 & - \\
SPP\cite{wang2023unleashing} & 7.6 & 8.0 & 8.3 & 7.1 & - \\
EvoAgent\cite{yuan2024evoagent} & 7.1 & 7.9 & 8.8 & 6.8 & - \\
ADAS\cite{hu2024automated} & 7.3 & 7.9 & 7.8 & 7.1 & 6.6 \\
    \midrule
\textbf{SwarmAgentic} & \textbf{8.5} & \textbf{8.3} & \textbf{9.0} & \textbf{7.5} &  \textbf{7.8}\\
\bottomrule
\end{tabular}
}
\caption{Performance on Creative Writing when transferring the best agentic system discovered by GPT-4o-mini to other LLMs. SwarmAgentic consistently outperforms all baselines across different LLMs, demonstrating strong cross-model transferability. Details of the best-discovered system are provided in Appendix \ref{app: Discovered Agentic System}. \textbf{*} indicates results where the agent is both trained on Gemini-1.5-flash~\cite{subramanya2024gemini} and tested on Gemini-1.5-Pro.}
\vspace{-5pt}
\label{tab:diff_llm}
\end{table*}

\subsection{Results} 
Tab. \ref{tab:tp-result} and \ref{tab:natural_plan_creative} report results across all tasks. Best agentic systems discovered by SwarmAgentic for each task are provided in Appendix \ref{app: Discovered Agentic System}.

\paragraph{SwarmAgentic achieves strong gains in open-ended, structurally unconstrained tasks.}
SwarmAgentic consistently outperforms all baselines on TP, NP, and CW—achieving a 261.8\% gain over ADAS on TP, leading all subtasks in NP, and generating more coherent outputs in CW. While prior frameworks rely on varying degrees of task-specific priors, SwarmAgentic operates solely based on a task description and an objective function. These results highlight the effectiveness of full autonomy in real-world tasks where static templates fall short. This underscores its generality across diverse tasks without hand-crafted assumptions.

\paragraph{Full automation remains effective in structured, template-compatible tasks.}
In MGSM, a math reasoning task with minimal structural variability, SwarmAgentic still achieves the best score. This demonstrates strong generalization and confirms that autonomy does not trade off performance even when predefined logic suffices.

\paragraph{SwarmAgentic surpasses both manual and automatic baselines through unified autonomy.}
Direct, CoT, and Self-Refine rely on fixed workflows, lacking adaptive structure. SPP, EvoAgent, and ADAS offer partial automation, but fall short of full autonomy: SPP depends on rigid persona templates, EvoAgent mutates fixed agent scaffolds, and ADAS initiates its search from hand-crafted seed agents. In contrast, SwarmAgentic constructs agent functionalities, behaviors, and collaboration strategies from scratch and jointly optimizes them with interpretable, feedback-driven updates, enabling scalable, task-specific agentic systems.

\begin{table}[htb]
\centering
\scalebox{0.76}{
\begin{tabular}{lcc}
     \toprule
     \textbf{Methods} & \textbf{Score} & \textbf{$\Delta$} \\ 
     \midrule
     Direct & 6.2  & 0\%\\
     \midrule
     \rowcolor{gray!30}
     \multicolumn{3}{c}{\textit{Different Iteration Count}}\\
     $\text{SwarmAgentic}_{\text{(3,1)}}$ & 5.9 & -4.8\% \\
     $\text{SwarmAgentic}_{\text{(3,5)}}$ & 6.4 & +3.2\%\\
     $\text{SwarmAgentic}_{\text{(3,10)}}$ & 7.0 & +12.9\%\\
     \midrule
     \rowcolor{gray!30}
     \multicolumn{3}{c}{\textit{Different Particle Count}}\\
     $\text{SwarmAgentic}_{\text{(1,5)}}$ & 6.3 & +1.6\%\\
     $\text{SwarmAgentic}_{\text{(3,5)}}$ & 6.7 &+8.1\%\\
     $\text{SwarmAgentic}_{\text{(5,5)}}$ & 6.9 &+11.3\%\\
     \midrule
     \rowcolor{gray!30}
     \multicolumn{3}{c}{\textit{Different Design Settings}}\\ 
     $\text{SwarmAgentic}_{\text{(5,10)}}$ \textit{\small w/o Collab. Struc. Reconfig.} & 6.7 & +8.1\%\\
     $\text{SwarmAgentic}_{\text{(5,10)}}$ \textit{\small w/o Agent-Level Adapt.} & 7.3 & +17.7\%\\ 
     $\text{SwarmAgentic}_{\text{(5,10)}}$ \textit{\small w/o Failure-Driven Adjust.} & 8.4 & +35.5\%\\
     $\text{SwarmAgentic}_{\text{(5,10)}}$ & 8.8&+41.9\%\\
     \bottomrule
\end{tabular}
}
\caption{Ablation Study on Creative Writing, evaluating the impact of key components and hyperparameters in SwarmAgentic. Removing Failure-Driven Adjustments, Agent-Level Adaptation, or Collaborative Structures Reconfiguration degrades performance, confirming their importance. Increasing iteration counts and particle counts improves performance, highlighting the benefits of iterative refinement and broader exploration. $\Delta$ indicates the differences compared with Direct.}
\vspace{-5pt}
\label{tab: AS-result}
\end{table}

\section{Analysis}
\subsection{Cross-Model Transferability Analysis}
We first optimize the agentic system using GPT-4o-mini and transfer the discovered system to other LLMs to test whether the system found with one model generalizes to others. As shown in Tab. \ref{tab:diff_llm}, the transferred SwarmAgentic system consistently outperforms all baselines, demonstrating strong cross-model generalizability. Notably, when SwarmAgentic is directly optimized and evaluated on Gemini-1.5-Pro (Gemini-1.5*), the performance further improves, indicating that model-specific optimization can yield additional gains. These results suggest that while SwarmAgentic systems exhibit robust transferability across foundation models, tailoring the optimization to the target LLM remains beneficial for achieving optimal results.

\subsection{Ablation Study}
We assess the impact of key components in SwarmAgentic, along with the effects of varying iteration counts and particle counts. A comprehensive analysis is conducted on 20 instances of the CW task, with results in Tab. \ref{tab: AS-result}.

\paragraph{Component Analysis.}
To analyze the optimization dynamics of SwarmAgentic, we assess the impact of its three core mechanisms: Failure-Driven Adjustments, Agent-Level Adaptation, and Collaborative Structures Reconfiguration. As shown in Tab. \ref{tab: AS-result}, removing failure-driven adjustments allows errors to persist across iterations, significantly impairing performance. Disabling agent-level adaptation restricts role flexibility, reducing system adaptability. Excluding collaborative structures reconfiguration disrupts task sequencing and dependency management, leading to execution inefficiencies.

\paragraph{Impact of Iteration and Particle Count.}
Tab.~\ref{tab: AS-result} shows that increasing either training iterations or particle count improves performance. More iterations enable progressive refinement through structured feedback, while a larger particle set enhances exploration, yielding up to +11.3\% improvement over Direct. These results highlight the benefits of both iterative optimization and population diversity in generating high-quality agentic systems.

\begin{figure}
    \centering
    \includegraphics[width=0.9\linewidth]{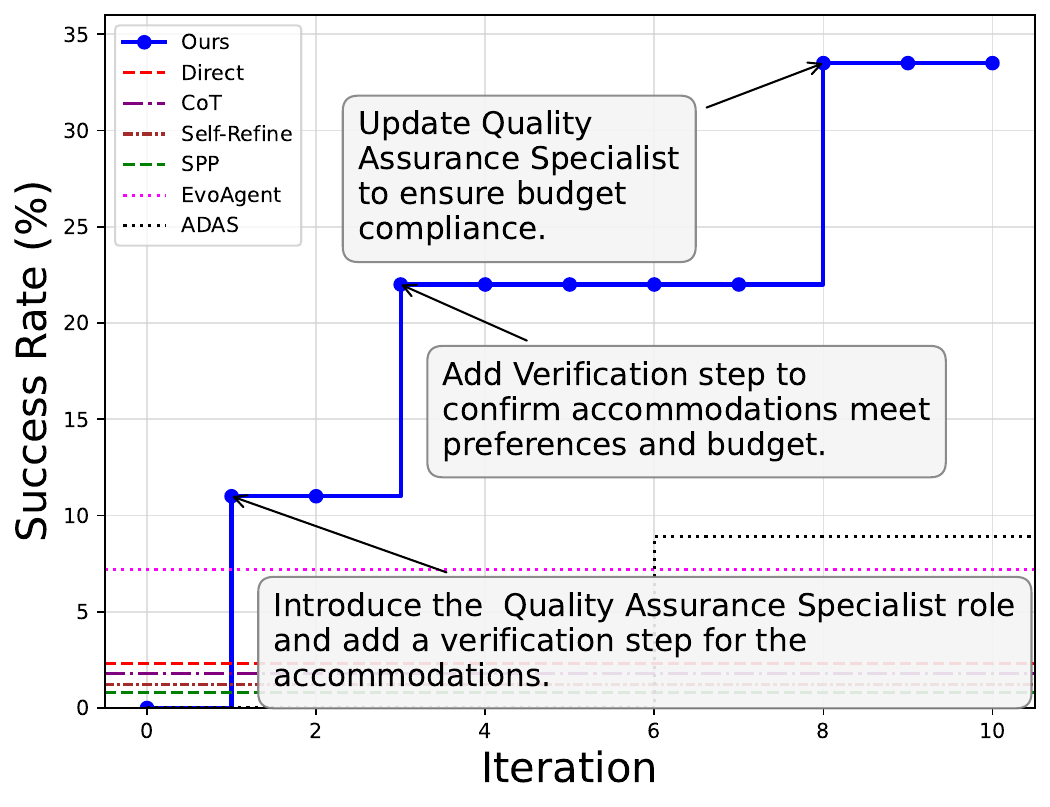}
    \caption{Search trajectory of SwarmAgentic on TravelPlanner. The Success Rate (SR) improves iteratively as specialized agents are introduced to refine constraint handling and enhance plan feasibility.}
    \label{fig:traj_new}
\vspace{-10pt}
\end{figure}

\subsection{Case Study: Search Trajectory on TP}
Fig.~\ref{fig:traj_new} illustrates the iterative optimization of SwarmAgentic on TP, refining both agent sets and collaborative structure. The process begins with introducing a Quality Assurance Specialist and a verification step for accommodations, boosting SR to 11\%. Adding a dedicated verifier to check budget and preference alignment raises performance to 22\%. Finally, the Quality Assurance Specialist is updated to explicitly enforce budget compliance, achieving a 33\% SR and surpassing all baselines. While the figure highlights agent evolution, collaborative structure optimization occurs in parallel, reconfiguring task dependencies and execution order to enhance coordination. See Appendix~\ref{app: Optimization Mechanism Illustration} for step-by-step illustrations of this evolution process, and Appendix~\ref{app: Comparison with ADAS-Discovered Agentic Systems} for representative agentic systems found by SwarmAgentic and ADAS.

\section{Conclusion}
We proposed SwarmAgentic, a language-driven PSO framework that enables fully automated, self-optimizing agentic systems. By integrating LLM-guided optimization, our method refines agent sets and collaborative structures dynamically, overcoming the rigidity of existing approaches. Extensive experiments on complex, real-world tasks show superior adaptability, constraint satisfaction, and coordination. SwarmAgentic bridges swarm intelligence and autonomous agent evolution, paving the way for scalable, self-optimizing agentic systems.

\section*{Limitations}
SwarmAgentic is designed for the automated construction of agentic systems in settings that lack predefined structural assumptions. While this promotes generalization to open-ended tasks, the framework does not incorporate inductive priors, such as domain-specific templates, that could help accelerate convergence in more structured environments. Incorporating such priors via language-driven initialization or constraint-guided search represents a promising direction for future work, offering a trade-off between structure-guided efficiency and open-ended flexibility.

Despite its effectiveness in automated agentic system generation, SwarmAgentic inherits several limitations intrinsic to LLMs, particularly in factual reliability and grounded interaction. As SwarmAgentic relies on LLMs for structured reasoning and decision-making, it remains susceptible to hallucinations—outputs that appear plausible but are factually incorrect. These errors can propagate through optimization cycles, compromising agent configurations and coordination structures. While iterative refinement helps alleviate such issues, persistent inaccuracies may necessitate the integration of external knowledge sources. Additionally, operating purely in a text-based environment, SwarmAgentic lacks perception and action capabilities in real-world contexts. In contrast to embodied systems, it cannot process multimodal inputs or interact with physical environments, which limits its applicability in dynamic, sensor-rich scenarios. Future extensions could explore integration with multimodal models or embodied agents to bridge this gap.

\bibliography{custom}

\newpage
\clearpage
\appendix
\onecolumn
\tableofcontents

\newpage

\section{Agentic System Generation Autonomy}
\label{app: Agentic Autonomy Evaluation Framework}
\subsection{Defining Autonomy: Three Core Properties}
\label{app: Defining Agentic Autonomy: Three Core Properties}
We define three core properties to evaluate the level of autonomy in agentic system generation. These properties are mutually exclusive and collectively reflect the system’s ability to construct, adapt, and scale agent-based solutions.

\begin{itemize}
  \item \textbf{From-Scratch Agent Generation} requires that the framework must dynamically synthesize complete agent instances—including their roles, decision logic, and internal structure—without relying on predefined functional modules, such as hard-coded operators, or task-specific behaviors. Minimal task-agnostic scaffolding (e.g., I/O wrappers or abstract interface definitions) may be reused, but all task-specific reasoning strategies, coordination flows, and behavioral compositions must be newly generated based on the task context. This capability is essential for real-world, open-ended tasks involving high-level planning, system-level coordination, and creative reasoning, where agent functionalities and coordination patterns must be automatically derived from the task description and objective function. Manual design or fixed generation pipelines impose structural priors that hinder adaptability and prevent the system from generalizing to novel or diverse scenarios.

\item \textbf{Self-Optimizing Agent Functionality} indicates whether an agent’s internal logic, such as its role, responsibility, or execution policy, can be automatically refined by the system itself, during execution or across iterations, in response to feedback or performance signals, without manual intervention. This dynamic adaptation must target the agent’s own behavior (not merely global workflow wiring) and go beyond a fixed, static prompt.  This is particularly important in exploratory tasks where agents often face ambiguous goals or unexpected failures. Without self-adjustment, the system would rely on brittle static prompts and require external corrections, undermining its autonomy and scalability.

\item \textbf{Self-Optimizing Agent Collaboration} indicates whether the framework can autonomously reconfigure collaborative structures, including task sequencing, dependency refinement, inter-agent coordination, and the addition or removal of execution steps. This supports dynamic restructuring of how agents interact to improve efficiency and adaptability. Effective collaboration in open-ended multi-agent settings demands flexibility: task decomposition, role delegation, and information flow often need to be revised mid-execution. Without the ability to restructure inter-agent workflows, the system cannot recover from coordination failures or adapt to emergent task constraints.
  
\end{itemize}

\subsection{Evaluation of Existing Agentic Frameworks}
\label{app: Autonomy Evaluation of Existing Agentic Frameworks}
We evaluate each baseline framework against these autonomy properties defined above. Below we provide justification for each binary assignment in Tab. \ref{table:benchmark-comparison}.

\begin{itemize}
\item SPP~\cite{wang2023unleashing} does not satisfy \textbf{From-Scratch Agent Generation}: it relies on a hard-coded multi-persona prompt scaffold that prescribes the three-stage pattern (persona identification → brainstorm → revision) and embeds two hand-crafted examples. The agents’ roles, dialogue order, and interaction protocol are therefore predefined rather than synthesised from the task. SPP also fails \textbf{Self-Optimizing Agent Functionality}: the underlying prompts and decision policies are frozen, so feedback only changes the answer text, not the agents’ own behaviour. It likewise fails \textbf{Self-Optimizing Agent Collaboration}, because the interaction pattern cannot be expanded, pruned, or reordered at run time.
  
  \item EvoAgent~\cite{yuan2024evoagent}  does not satisfy \textbf{From-Scratch Agent Generation}: evolution begins from a hand-written specialist agent supplied by MetaGPT, AutoGen, or a similar template, and merely mutates its roles, skills, and prompts, so the core logic is derived rather than synthesised directly from the task. It does satisfy \textbf{Self-Optimizing Agent Functionality}, as LLM-guided mutation plus fitness evaluation iteratively refines each agent’s internal behaviour. However, it fails \textbf{Self-Optimizing Agent Collaboration}: the interaction topology is fixed by the underlying framework—evolution can modify individuals but cannot reorder tasks, alter message routing, or create new coordination flows.

  \item AgentSquare~\cite{shang2024agentsquare} does not satisfy \textbf{From-Scratch Agent Generation}: search starts from a fixed library of four standardised module types—planning, reasoning, tool-use, and memory—extracted from sixteen existing agents. New agents are only recombinations or mutations of these predefined modules, so core behaviour is not synthesised solely from the task description. It satisfies \textbf{Self-Optimizing Agent Functionality}, since each module can be mutated or rewritten by the LLM and retained or discarded based on performance, allowing an agent’s internal logic to evolve across iterations. It fails \textbf{Self-Optimizing Agent Collaboration}: the framework optimizes a single-agent modular architecture and never reconfigures multi-agent interaction patterns or execution topology. 
  
 \item AutoAgents~\cite{chen2023autoagents} does not satisfy \textbf{From-Scratch Agent Generation}: the framework is hard-wired with four manager roles—\textit{Planner}, \textit{Agent-Observer}, \textit{Plan-Observer}, and a run-time \textit{Action-Observer}. These modules embed planning, evaluation, and dispatch logic, exceeding the allowance for minimal task-agnostic scaffolding and anchoring core reasoning to a preset template rather than synthesizing it solely from the task description. These human-designed interventions limit their scalability and functionality~\cite{yuan2024evoagent}. Specifically, AutoAgents relies on these four predefined manager roles, and all agent generation and collaboration processes must revolve around them. The agent team structure and execution plan are not freely synthesized solely from the task but are constrained within a fixed template. This restricts the system’s flexibility and dynamic generation capability in adapting to complex and variable tasks. For example, it cannot effectively handle highly open-ended tasks like TravelPlanner that require dynamic multi-role and complex constraint coordination. It satisfies \textbf{Self-Optimizing Agent Functionality}: each task-specific expert executes a \textsc{Think} $\!\to\!$ \textsc{Plan} $\!\to\!$\textsc{Act} $\!\to\!$ \textsc{Reflect} loop that automatically rewrites its own prompt, plan, and memory in response to feedback. It also satisfies \textbf{Self-Optimizing Agent Collaboration}: the planner–observer dialogue can add or remove experts and resequence steps, while the \textit{Action-Observer} dynamically adjusts the plan during execution.

\item AFlow~\cite{zhang2024aflow} fails \textbf{From-Scratch Agent Generation}: it assembles workflows from a fixed palette of hard-coded operators (\texttt{Generate}, \texttt{Revise}, \texttt{Ensemble}, \texttt{Test}), so task-specific logic is selected rather than newly synthesized. It satisfies \textbf{Self-Optimizing Agent Functionality}, as execution feedback triggers automatic prompt edits, control-flow tweaks, and operator replacement without human input. It also satisfies \textbf{Self-Optimizing Agent Collaboration}, because the MCTS search can dynamically reorder tasks, add or prune branches, and revise coordination strategies.

\item Agent Symbolic Learning~\cite{zhou2024symbolic} does not satisfy \textbf{From-Scratch Agent Generation}, because it starts from a manually crafted pipeline inherited from prior work~\cite{zhou2023agents} rather than synthesising roles directly from the task description. It does meet \textbf{Self-Optimizing Agent Functionality}: each node’s prompt and tool usage are refined via symbolic gradients driven by language loss. The system also satisfies \textbf{Self-Optimizing Agent Collaboration}, since the pipeline optimizer can add, delete or move nodes to restructure coordination.

  \item ADAS~\cite{hu2024automated} fails \textbf{From-Scratch Agent Generation}: the search starts from seven hand-written seed agents, so new agents are mutated variants of these seeds rather than being created solely from the task description. It satisfies \textbf{Self-Optimizing Agent Functionality}, as the meta-agent repeatedly rewrites each candidate’s code, prompts, and tool calls using performance feedback, preserving only the best variants. It satisfies \textbf{Self-Optimizing Agent Collaboration}, since the meta-agent can insert or remove internal roles and reorder their interactions, letting coordination structures evolve across iterations. 
\end{itemize}

\section{Comparison with MODEL SWARMS: From Model Fusion to Agentic System Generation}
\label{app: Comparison with MODEL SWARMS: From Model Fusion to Agentic System Generation}
MODEL SWARMS~\cite{feng2024model} is a collaborative optimization framework that adapts pretrained LLM experts by searching in the model weight or token probability space. It applies particle swarm optimization (PSO) to iteratively interpolate and update a pool of existing models, guided by a task-specific utility function. The goal is to discover a single adapted model that performs well under limited data conditions, without requiring fine-tuning or strong assumptions about expert composition.

Despite sharing high-level inspiration from swarm intelligence, our approach differs fundamentally from MODEL SWARMS in objective, search space, optimization strategy, and output structure. While MODEL SWARMS optimizes model parameters within a fixed expert pool, our method explores a language-based agentic system design space. We construct executable multi-agent systems from scratch—each comprising dynamic roles, internal logic, tool usage, and coordination strategies—based solely on task descriptions. Additionally, whereas MODEL SWARMS relies on interpolation and performance-based selection, we introduce a Failure-Aware Velocity Update mechanism that performs symbolic, LLM-guided rewriting of agent functionalities and collaboration flows. Finally, the outputs are categorically distinct: MODEL SWARMS produces a single, opaque model optimized for static evaluation, while our framework generates a modular, interpretable agentic system capable of reasoning, adapting, and evolving in complex, dynamic environments. This marks a paradigm shift from model fusion to full-system generation.

\section{Experimental Setup}
\label{app: Experimental Details}
\subsection{Dataset Statistics and Evaluation}
\label{app: Dataset Statistics and Evaluation}

\paragraph{MGSM} Following~\cite{hu2024automated}, we sample 128 training and 800 test questions. 

\paragraph{Creative Writing} We use all 100 tasks, reserving the first 5 for training and the remaining 95 for evaluation.

\paragraph{Natural Plan} We train on a difficulty-balanced subset of the Natural Plan dataset: one example per difficulty level—cities-to-visit $N\in[3,10]$ for Trip Planning, friends-to-meet $N\in[1,10]$ for Meeting Planning, and Calendar Scheduling with (i) one-day schedules ($N\in[3,7]$ meetings) and (ii) two-day schedules ($D\in[1,5]$ days apart). This results in 8 Trip Planning and 10 Meeting and Calendar Scheduling training examples. Evaluation is conducted on a held-out validation set comprising 10\% of the full dataset, sampled with the same difficulty distribution and disjoint from the training data to avoid leakage.

\paragraph{TravelPlanner} We follow the setup in~\cite{yuan2024evoagent} and evaluate on 180 user queries. For training, we use 9 representative queries from the original TravelPlanner training set, selected to match the difficulty distribution of the validation set.

\paragraph{Evaluation Metrics}
For all tasks, we follow the evaluation metrics established in the original setting. (1) TP is assessed based on delivery rate, commonsense constraint pass rate, hard constraint pass rate, and final pass rate, with micro and macro strategies providing a detailed analysis of constraint satisfaction; (2) NP employs an exact match score, where generated plans are compared against ground truth using regex-based parsing to extract key details; (3) CW is evaluated using LLM with a zero-shot prompt, assigning scalar scores (1-10) and averaging five samples per output to enhance reliability; (4) MGSM employs an exact match score, where the generated integer answer is compared directly with the reference answer for correctness.

\subsection{Baselines and Configurations}
\label{app: Baseline Implementations and Configurations}

We detail the setup for all baselines to ensure a fair and representative comparison. For each method, we follow the official implementation and apply task-specific adaptations where required, consistent with the original design intent.

\begin{enumerate}
\item \textbf{Direct} The LLM answers the input directly without intermediate reasoning or feedback.

\item \textbf{CoT}~\cite{wei2022chain}. The LLM is prompted to reason step by step before producing a final answer.

\item \textbf{Self-Refine}~\cite{madaan2024self}. We adopt the iterative refinement pipeline proposed in the original paper, using the official codebase and settings.

\item \textbf{SPP}~\cite{wang2023unleashing}. We follow the structured persona prompting format from the original paper. The persona pool and dialogue structure are fixed across tasks, reflecting its hard-coded multi-agent interaction template.

\item \textbf{EvoAgent}~\cite{yuan2024evoagent}. We adopt the official mutation strategies and role initialization schemes from the released implementation.

\item \textbf{ADAS}~\cite{hu2024automated}. We employ the full Meta Agent Search framework, including 7 pre-written seed agents and meta-agent rewriting policies. Following the original setup, we update task-specific information (e.g., constraints and formats) in the meta-agent prompt to reflect each domain. Additionally, we adapted the role-based methods from the initial library to better fit each task.
\end{enumerate}

\textbf{Prompt Adaptation} For all methods, we made necessary prompt word adjustments to fit each task (e.g., "writing result" instead of "answer" for Creative Writing) while preserving each method's logic. No additional search or adaptation beyond the original algorithm was performed.

\section{SwarmAgentic Implementation}
\label{app: Implementation Details}

\subsection{Basic Structure of Agentic System}
\label{app: Basic Structure of Agentic System}
We implement a modular framework for role-based multi-agent collaboration. The system defines structured classes for dynamically instantiating callable functions, parsing inputs, and orchestrating multi-role execution. The \texttt{Role} class serves as a structural placeholder for role-specific behavior, execution policies, and responsibilities, which are dynamically instantiated and optimized via LLM-guided search during the PSO process. The \texttt{Team} class manages inter-agent coordination and information flow. This architecture supports flexible task delegation and compositional control, and is designed for automated agentic system generation and refinement. This framework forms the structural backbone of SwarmAgentic, enabling dynamic agent instantiation and coordination during the PSO-driven search process.

\lstset{
  basicstyle=\ttfamily\small,
  keywordstyle=\color{blue},
  commentstyle=\color{gray},
  stringstyle=\color{red},
  numberstyle=\tiny,
  stepnumber=1,
  numbersep=5pt,
  backgroundcolor=\color{white},
  showspaces=false,
  showstringspaces=false,
  showtabs=false,
  frame=single,
  breaklines=true
}
\begin{lstlisting}[language=Python]
def set_forward(next_solution):
    """
    Dynamically creates and returns a callable Python object defined by the input code string.

    Args:
        next_solution (str): A string containing valid Python code that defines a function or a callable object.

    Returns:
        Callable: The function or callable object generated from the provided code.
    """
    ...
    return func

class Role():
    """
    Base class representing a role within an agentic system.

    Attributes:
        name (str): Name of the agent.
        responsibility (str): Description of the agent's responsibility.
        policy (str): Operational policy or behavioral guideline for the agent.
        llm (Any): Language model instance used for generating responses.
        message (Any): Object that stores the agent's most recent communication.
    """

    def __init__(self, role: dict, llm) -> None:
        ...
    
    def parse_inputs(self, inputs: List) -> str:
         """
        Constructs a task prompt based on the provided inputs.

        Args:
            inputs (List): A list of inputs, typically including the task and outputs from other agents.

        Returns:
            Tuple[str, str]: A tuple containing the current task instance and combined outputs from other agents.
        """
        ...
        return task_instance, others_outputs
        
    def response(self, task_instance, others_outputs, output):                    
        """
        Generates the agent's response using LLM.
        Args:
            task_instance (str): The current task or instruction for this agent.
            others_outputs (str): Outputs or messages received from other agents.
            output (str): Desired output format or specification.

        Returns:
            str: The final response generated by the agent.
        """  
        return self.message.content  

    def __call__(self, inputs, output):
        """
        Executes the agent's full decision-making process: input parsing, response generation, and return.

        Args:
            inputs (List): List of inputs, including task and other agents' outputs.
            output (str): Output format specification.

        Returns:
            str: The response generated by the agent.
        """
        task_instance, others_outputs = self.parse_inputs(inputs)
        return self.response(task_instance, others_outputs, output)

class Team():
    """
    class for a team, which consists of multiple agents and a workflow about how they interact with each other. A particle consists of a team, composed of multiple interacting agents defined by a workflow, and the executable code generated by LLM_write_forward.

    Attributes:
    - llm: LLM model to be used
    - roles: List of agents in the team
    - workflow: Workflow of the team
    - task: Task to be solved by the team
    - message_pool: Message pool for the team
    """    
    def __init__(self, llm, logger) -> None:
        ...
    def call(self, required_role: str, inputs: List = [], output: str = ""):
        """call the role with the required agent name. The inputs are the outputs from other agents.

        Args:
            required_role (str): name of the required agent.
            inputs (List, optional): inputs for the agent. Defaults to [].
            output (str, optional): output requirements for the agent. Defaults to "".

        Returns:
            response: response of the role.
        """        
        ...
        return responses
\end{lstlisting}

\subsection{Pseudo Code for SwarmAgentic}
\label{app: Pseudo Code for SwarmAgentic}

\begin{algorithm}[H]
  \centering
  \caption{Agentic System Search with Particle Swarm Optimization}
  \begin{algorithmic}[1]               

  \REQUIRE LLM temperatures $\{\text{temp}_i\}_{i=1}^n$, fitness function
           $J : x \!\to\! \mathbb{R}$;  
           system initialization function $\text{LLM}_{\text{init\_team}}$;
           system code-generation function $\text{LLM}_{\text{write\_forward}}$;
           performance evaluation function $\text{LLM}_{\text{eval}}$;
           system flaw identification function $\text{LLM}_{\text{flaw}}$;
           velocity initialization function $\text{LLM}_{\text{init\_vel}}$;
           failure identification function  $\text{LLM}_{\text{identify\_fail}}$;
           learning from failure function $\text{LLM}_{\text{fail}}$;
           global best guidance function $\text{LLM}_{\text{glob}}$;
           personal best guidance function $\text{LLM}_{\text{pers}}$;
           velocity update function $\text{LLM}_{\text{vel}}$;
           position update function $\text{LLM}_{\text{pos}}$;
           
           swarm size $N$, max iteration $T$
  \STATE \textbf{Input:}  dataset for training $D$
  \STATE \textbf{Output:} global best checkpoint $g$

  \STATE \textit{// Initialize search}
  \FOR{$i \gets 1$ \TO $N$}
      \STATE \textit{// LLM\textsubscript{write\_forward} enables automatic code execution}
      \STATE \textbf{Initialize position:} $x_i^{(0)} \gets \text{LLM}_{\text{init\_team}}(\text{temp}_i)$
      \STATE \textbf{Initialize fitness:}  $j_i^{(0)} \gets J(x_i^{(0)},D)$
      \STATE \textbf{Identify Problem:} $p_{i}^{(0)} \gets \text{LLM}_{\text{eval}}(x_i^{(0)}, j_i^{(0)})$
      \STATE \textbf{Refection Summarization:} $f_{i}^{(1)} \gets \text{LLM}_{\text{flaw}}(x_i^{(0)}, p_{i}^{(0)})$
      \STATE \textbf{Initialize velocity:} $v_i^{(1)} \gets \text{LLM}_{\text{init\_vel}}(x_i^{(0)}, f_{i}^{(1)}))$
      \STATE \textbf{Initialize personal best:} $p_i^* \gets x_i^{(0)}$,\;
             $j_{p,i} \gets j_i^{(0)}$
      \STATE \textbf{Update Position:} $x_i^{(1)} \gets \text{LLM}_{\text{pos}}(x_i^{(0)}, v_i^{(1)})$\hfill 
  \ENDFOR
  \STATE \textbf{Initialize global best}: $g \gets \arg\max_{i} j_{p,i}^{(0)},\; f_g \gets \max_{i} j_{p,i}^{(0)}$

  \STATE \textit{// Start search}
  \FOR{$t\gets 1$ \TO $T$}
      \FORALL{$i = 1,\dots,N$ \textbf{(in parallel)}}
          \STATE \textbf{Execution:}
          \STATE \quad \textbf{Update Fitness:} $j_i^{(t)} \gets J(x_i^{(t)},D)$
          \STATE \quad\textbf{Identify Problem:} $p_{i}^{(t)} \gets \text{LLM}_{\text{eval}}(x_i^{(t)}, j_i^{(t)})$
          \STATE \quad\textbf{Refection Summarization:} $f_{i}^{(t+1)} \gets \text{LLM}_{\text{flaw}}(x_i^{(t)}, p_{i}^{(t)})$

          \textbf{Update Global Best:}\IF{$j_i^{(t+1)} > j_g$}
              \STATE $g \gets x_i^{(t+1)}$;\; $j_g \gets j_i^{(t+1)}$
          \ENDIF
          
          \textbf{Update Personal Best:}\IF{$j_i^{(t+1)} > j_p$}
              \STATE $p_i^* \gets x_i^{(t+1)}$;\; $j_{p,i} \gets j_i^{(t+1)}$
          \ENDIF
          \STATE \textbf{Update Velocity:}
          \STATE \quad \textit{// LLM\textsubscript{identify\_fail} identify the previous failed adjustments}
          \STATE \quad $c_f r_f  F(v_i^{(t)}) = \text{LLM}_{\text{fail}}(v_i^{(t)}, f_i^{(t)}, f_i^{(t+1)})$\hfill $\triangleright$ Eq. \eqref{eq:faliure}
          \STATE \quad $c_p r_p (p_i^* - x_i^{(t)}) = \text{LLM}_{\text{pers}}(x_i^{(t)}, p_i^*, f_i^{(t+1)})$\hfill $\triangleright$ Eq. \eqref{eq:personal_best}
          \STATE \quad $c_g r_g (g - x_i^{(t)}) = \text{LLM}_{\text{glob}}(x_i^{(t)}, g, f_i^{(t+1)})$\hfill $\triangleright$ Eq. \eqref{eq:global_best}
          \STATE \quad $v_i^{(t+1)} = \text{LLM}_{\text{vel}}(c_f r_f F(v_i^{(t)}), c_p r_p (p_i^* - x_i^{(t)}), c_g r_g (g - x_i^{(t)}))$\hfill $\triangleright$ Eq. \eqref{eq:update_vel}
          \STATE \textbf{Update Position:} $x_i^{(t+1)} = \text{LLM}_{\text{pos}}(x_i^{(t)}, v_i^{(t+1)})$\hfill $\triangleright$ Eq. \eqref{eq:position_update}

      \ENDFOR
  \ENDFOR

  \RETURN $g$
  \end{algorithmic}
  \label{alg:pso}
\end{algorithm}

\subsection{Prompt Repository}
\label{app: Prompt Repository}
We employ the following prompts to achieve the automated generation of agentic systems with PSO. Specifically, we use $\mathrm{LLM}_{\text{init\_team}}$ to initialize both the roles and the team for each particle at the start of the process, ensuring consistency in team composition and task allocation. $\mathrm{LLM}_{\text{write\_forward}}$ is then used to generate the corresponding code based on the initialized roles and the given workflow, enabling forward progression of each particle's function. To identify problems in the responses, we employ performance evaluation $\mathrm{LLM}_{\text{eval}}$, which analyzes the workflow and task execution to reveal underlying issues and explain their root causes in relation to the intended process. Once a problem is identified, flaw identification $\mathrm{LLM}_{\text{flaw}}$ is applied to trace it back to underlying issues in the role or team configuration. This step helps uncover structural or logical flaws that may hinder performance.

Next, we initialize the velocity of each particle $\mathrm{LLM}_{\text{init\_vel}}$, considering the current team composition and the identified design flaws. This initialization provides direction and momentum for adjustment in future iterations. We then examine the failed adjustments $\mathrm{LLM}_{\text{identify\_fail}}$ from the previous iteration using a specialized prompt designed to extract and document ineffective changes. The Learning from Failure prompt $\mathrm{LLM}_{\text{fail}}$ is used to suggest improved strategies, leveraging past failures to guide more effective future adjustments. To complement this, we use additional prompts to discover meaningful adjustments inspired by both the global best team $\mathrm{LLM}_{\text{glob}}$ and the personal best team $\mathrm{LLM}_{\text{pers}}$, promoting convergence towards optimal configurations. Velocity is updated $\mathrm{LLM}_{\text{vel}}$ by integrating suggestions from global best guidance, personal best guidance, and failure-driven learning. This multi-source adjustment balances exploration and exploitation. Finally, the team configuration is updated $\mathrm{LLM}_{\text{pos}}$ according to the plan generated during the velocity update phase, completing one full iteration of the optimization cycle and preparing for the next.

\begin{tcolorbox}[colback=white, colframe=black, title=Prompt Template for Agents, width=\linewidth, halign=left, enhanced jigsaw, breakable]
\begin{lstlisting}[language=TeX, basicstyle=\ttfamily\footnotesize, numbers=none, xleftmargin=0pt, frame=none]
ROLE_PROMPT = '''You are {name}. You are working in a team solving the following specific task:
<task instance>
{instance}
</task instance>

You are also provided with helpful information from other team members: 
<helpful information>
{information}
</helpful information>

# Instruction
Based on the <task instance> and <helpful information>, your responsibility is: {responsibility}
Please follow the instructions step by step to give an answer: 
<instruction>
{policy}
</instruction>

# Output Guidance
Your answer only needs to include: {output}
Think step by step and limit your answer to 100 words.
'''
\end{lstlisting}
\end{tcolorbox}

\begin{tcolorbox}[colback=white, colframe=black, title=Prompt for Agentic System Initialization $\mathrm{LLM}_{\text{init\_team}}$, width=\linewidth, halign=left, enhanced jigsaw, breakable]
\begin{lstlisting}[language=TeX, basicstyle=\ttfamily\footnotesize, numbers=none, xleftmargin=0pt, frame=none]
You are an expert in designing a highly efficient, specialized, and collaborative multi-agent team for a specific task.

**Requirements:**
- The team must break down the task into highly specialized, modular roles.
- Each role should have a focused domain of responsibility, handling only one primary aspect of the task.
- The information flow must be strictly modular, with each step primarily receiving structured input from the outputs of previous steps. Steps can refer to the initial task definition implicitly as needed, but it should not be treated as a direct input for workflow dependencies.
- Each step's output must be structured and usable as a direct input for subsequent steps, creating a clear, step-by-step workflow.
- Each step can only be assigned to a single role and cannot involve multiple roles simultaneously.
- The resulting team structure should allow for easy scalability and clarity, ensuring that each module can be independently optimized or replaced without affecting other parts of the system.

**Deliverables:**
1. Define Each Role:
   - Name: A clear and descriptive title.
   - Responsibility: A narrowly focused set of tasks aligned with that domain.
   - Policy: Specific operational guidelines for fulfilling these tasks.
2. Collaboration Structure:
   - Clearly outline how roles interact and pass information to one another.
   - Ensure that information flows from one role to another in a well-defined manner. Each role should clearly know which role's output it relies on, if any. If there is no upstream role, it operates independently (with no input).
3. Sequential Workflow:
   - Illustrate a concrete workflow from start to finish.
   - For each step:
      * Specify the single role responsible for that step.
      * Define its input, which must come from previous roles' outputs or be empty.
      * Define its output, which will be used as input for subsequent steps.
   - Ensure there is a designated role at the end to integrate all components into the final deliverable.

Now, giving the following task: {task}

Please design a detailed multi-agent collaborative team that could efficiently solve the task.
\end{lstlisting}
\end{tcolorbox}

\begin{tcolorbox}[colback=white, colframe=black, title=
Prompt for System Compilation 
$\mathrm{LLM}_{\text{write\_forward}}$
, width=\linewidth, halign=left, enhanced jigsaw, breakable]
\begin{lstlisting}[language=TeX, basicstyle=\ttfamily\footnotesize, numbers=none, xleftmargin=0pt, frame=none]
You are an expert Python programmer. You are tasked with writing 
a function to organize available roles to solve a specific task.
{function description}

You are provided with the following available roles. Each role 
can solve a subtask of the complex task:
<available roles>
{roles}
</available roles>

You are also given the workflow of these roles: 
<workflow>
{workflow}
</workflow>

Your job is to design the function that represents how the roles 
will work together to solve the task. 
Use these guidelines when generating the function:
- ALWAYS use **role_response = team.call(role_name: str, inputs: List, output: str)** to call a role. This will give inputs and required output instructions to the role and return the role's response. 
    * role_name: The name of the role to call in this step. You can only call roles in the current team. MUST NOT call a non-existent role from the available roles.
    * inputs: List of the outputs produced by one or more roles in the previous steps. 
    * output: What output is expected from the role in this step? Must be enclosed in double quotation marks ("output").
- Use the provided workflow instruction as a guide for designing the function's structure.
- Create a well-organized function that represents how the roles will work together to solve the task efficiently. 
- MUST not make any assumptions in the code.
- Ensure that every variable declared in the function is utilized, with no unused or redundant variables.
- Ensure the created function is complete and correct to avoid runtime failures.

# Examples

Here is an example to help you design the function:
<examples>
{examples}
</examples>
\end{lstlisting}
\end{tcolorbox}

\begin{tcolorbox}[colback=white, colframe=black, title=
Prompt for Performance Evaluation  
$\mathrm{LLM}_{\text{eval}}$
, width=\linewidth, halign=left, enhanced jigsaw, breakable]
\begin{lstlisting}[language=TeX, basicstyle=\ttfamily\footnotesize, numbers=none, xleftmargin=0pt, frame=none]
You are an expert assistant. You are tasked with analyzing the given workflow to identify where issues occurred, leading to the problem. You must provide a detailed explanation of the cause of the error.

The team is solving the following tasks:
<task>
{task}
</task>

The roles are collaborative in the following workflow:
<workflow>
{workflow}
</workflow>

You are also provided with the problem in the team result:
<problem>
{evaluation}
</problem>

Please provide a detailed explanation of the root cause of the 
<problem> at the identified step(s) with by referencing the 
detail information of the <task>, while considering factors such 
as incorrect execution, missing information, or deviations from the intended process.
\end{lstlisting}
\end{tcolorbox}

\begin{tcolorbox}[colback=white, colframe=black, title=
Prompt for Flaw Identification 
$\mathrm{LLM}_{\text{flaw}}$
, width=\linewidth, halign=left, enhanced jigsaw, breakable]
\begin{lstlisting}[language=TeX, basicstyle=\ttfamily\footnotesize, numbers=none, xleftmargin=0pt, frame=none]
You are an expert assistant tasked with reflecting on feedback and indicating specific flaws in the current team.

Given the following feedback:
<feedback> 
{feedback}
</feedback>

The team to optimize is as follows, including its roles and collaborative workflow:
<current team>
{current team}
</current team>

# Instruction

Based on the <feedback>, identify the specific flaws in the roles or workflow steps that directly contributed to the <feedback>. The flaw should be within the following types:
    1. Missing Role: Were there missing roles in the team that left certain tasks inadequately addressed or overlooked? Clearly specify which role may be needed.
    2. Redundant Role: Were there redundant roles in the team that were unnecessary? Clearly indicate the specific role that is redundant.
    3. Role Policy Deficiency: If the policy of the role is sufficiently instructive, clear, and effective. Are there gaps, ambiguities, or contradictions in the policy that affect role performance? Clearly specify the name of the role.
    4. Missing Workflow Step: Were there missing steps in the workflow that left certain tasks inadequately addressed or overlooked? Clearly specify between which two steps the missing step should have occurred.
    5. Redundant Workflow Step: Were there redundant steps in the workflow that are unnecessary? Clearly indicate the specific role and the exact step number that is redundant.
    6. Insufficient Input: Were the inputs insufficient for the workflow steps? Assess if it includes all the necessary information needed to get the role's output with its responsibility effectively. Clearly specify the role responsible for the step and the exact step number where the input was insufficient.
    7. Inappropriate Output: Before identifying an output as inappropriate, verify whether the requested output falls within the role's scope of responsibility. If the requested output exceeds the role's responsibility, reassign the task to an existing role better suited for it or create a new role specifically responsible for the output if no such role exists. Only when the required output is within the role's responsibility and still incorrect, missing, or incomplete should it be classified as inappropriate output for that role. Clearly specify the role responsible for the step and the exact step number where the output was inappropriate.
\end{lstlisting}
\end{tcolorbox}

\begin{tcolorbox}[colback=white, colframe=black, title=
Prompt for Velocity Initialization 
$\mathrm{LLM}_{\text{init\_vel}}$
, width=\linewidth, halign=left, enhanced jigsaw, breakable]
\begin{lstlisting}[language=TeX, basicstyle=\ttfamily\footnotesize, numbers=none, xleftmargin=0pt, frame=none]
You are tasked with optimizing a multi-agent team setup to enhance its performance in solving a specific task.

The team to optimize is as follows, including its roles and collaborative workflow: 
<current team>
{current_team}
</current team>

However, the <current team>'s performance is insufficient and must be improved based on the following feedback:
<feedback> 
{feedback} 
</feedback>

# Instruction

Follow the instructions to generate your response:
- Use the following OPERATIONS to refine roles within the <current team>: 
    * Add Role: Introduce a new role when an existing subtask becomes overly complex or burdensome, requiring a specialized responsibility that cannot be integrated into current roles without disrupting their primary responsibilities. Define the role's:
        - Name: A clear name that reflects its specific responsibility.
        - Responsibility: Specific tasks or functions the role will handle.
        - Policy: Operational guidelines for fulfilling the role's duties.
    * Modify Role: Adjust the policy of an existing role for improved role execution, when the identified inefficiencies or gaps can be addressed through manageable refinements to its policy, ensuring the changes do not overburden the role and are within the scope of its responsibility.
    * Delete Role: Remove roles that are redundant, unnecessary, or conflict with the team's primary objectives.
- Use the following OPERATIONS to optimize the workflow of the <current team>:
    * Add Step: Add a new step if a gap exists in the workflow that hinders overall efficiency, coordination, or goal achievement. Ensure the new step does not duplicate the functions of existing steps and adds clear value to the process. Define the step's:
        - Role: The role responsible for acting in this step.
        - Input: The input for this step must be the output produced by one or more roles in previous steps.
        - Output: What output is expected from the role in this step?
    * Modify Input: Adjust the input of an existing workflow step to ensure that it comprehensively incorporates outputs from previous steps to support the current step.
    * Modify Output: Modify the output of an existing workflow step to ensure that it fully aligns with the expected deliverables of the step and supports the inputs of subsequent steps.
    * Delete Step: Delete a step if it has become redundant, no longer contributes to team goals, or overlaps with other steps in the workflow. Ensure the removal of the step does not impact other steps' efficiency or completeness in achieving objectives.
    * Re-order Steps: Re-order steps if their current sequence causes inefficiencies or coordination issues within the workflow. Ensure the new order improves logical flow without compromising the integrity or dependencies of other steps.
- For each identified flaw in <feedback>, apply the following steps:
    * Identified Flaw: Clearly outline the specific flaw identified in the <feedback> section.
    * Proposed Adjustment: Specify the exact OPERATIONS to address the **Identified Flaw**.
\end{lstlisting}
\end{tcolorbox}

\begin{tcolorbox}[colback=white, colframe=black, title=
Prompt for Failure Identification  
$\mathrm{LLM}_{\text{identify\_fail}}$
, width=\linewidth, halign=left, enhanced jigsaw, breakable]
\begin{lstlisting}[language=TeX, basicstyle=\ttfamily\footnotesize, numbers=none, xleftmargin=0pt, frame=none]
You are a strategic advisor focused on enhancing the team's performance. Your role is to carefully analyze the feedback provided and identify failed adjustments with the previous adjustment plan.

You are given the following feedback on areas for the team 
improvement:
<feedback> 
{feedback}
</feedback>
 
You are also provided with the previous adjustment plan, the measures taken to enhance team performance:
<previous adjustment plan>
{velocity}
</previous adjustment plan>

# Instruction
   
For each flaw in <feedback>, please apply the following steps:
1. Identified Flaw: 
    - Clearly outline the specific flaw identified in the <feedback> section.
2. Thought: 
    - Carefully think if there is any **Proposed Adjustment** in the <previous adjustment plan> section for the exact same **Identified Flaw**. 
3. Failed Adjustment: 
    - Based on your **Thought**, quote the exact **Proposed Adjustment** as described in <previous adjustment plan> if there is any **Proposed Adjustment** for the same kind of Identified Flaw in <previous adjustment plan>. Otherwise, say 'None' here.
\end{lstlisting}
\end{tcolorbox}

\begin{tcolorbox}[colback=white, colframe=black, title=
Prompt for Learning from Failure  
$\mathrm{LLM}_{\text{fail}}$
, width=\linewidth, halign=left, enhanced jigsaw, breakable]
\begin{lstlisting}[language=TeX, basicstyle=\ttfamily\footnotesize, numbers=none, xleftmargin=0pt, frame=none]
You are a strategic advisor focused on enhancing the team's performance. Your role is to carefully analyze the feedback provided and align team improvements with previous adjustment directions.

The team to optimize is as follows, including its roles and collaborative workflow: 
<current team>
{team}
</current team>

You are given the following feedback, including every "Identified Flaw" and its "Failed Adjustment":
<feedback> 
{feedback}
</feedback>

# Instruction

Follow the instructions to generate your response:
- Use the following OPERATIONS to refine roles within the <current team>: 
    * Add Role: Introduce a new role when an existing subtask becomes overly complex or burdensome, requiring a specialized responsibility that cannot be integrated into current roles without disrupting their primary responsibilities. Define the role's:
        - Name: A clear name that reflects its specific responsibility.
        - Responsibility: Specific tasks or functions the role will handle.
        - Policy: Operational guidelines for fulfilling the role's duties.
    * Modify Role: Adjust the policy of an existing role for improved role execution when the identified inefficiencies or gaps can be addressed through manageable refinements to its policy, ensuring the changes do not overburden the role and are within the scope of its responsibility.
    * Delete Role: Remove roles that are redundant, unnecessary, or conflict with the team's primary objectives.
- Use the following OPERATIONS to optimize the workflow of the <current team>:
    * Add Step: Add a new step if a gap exists in the workflow that hinders overall efficiency, coordination, or goal achievement. Ensure the new step does not duplicate the functions of existing steps and adds clear value to the process. Define the step's:
        - Role: The role responsible for acting in this step.
        - Input: The input for this step must be the output produced by one or more roles in previous steps.
        - Output: What output is expected from the role in this step?
    * Modify Input: Adjust the input of an existing workflow step to ensure that it comprehensively incorporates outputs from previous steps to support the current step.
    * Modify Output: Modify the output of an existing workflow step to ensure that it fully aligns with the expected deliverables of the step and supports the inputs of subsequent steps.
    * Delete Step: Delete a step if it has become redundant, no longer contributes to team goals, or overlaps with other steps in the workflow. Ensure the removal of the step does not impact other steps' efficiency or completeness in achieving objectives.
    * Re-order Steps: Re-order steps if their current sequence causes inefficiencies or coordination issues within the workflow. Ensure the new order improves logical flow without compromising the integrity or dependencies of other steps.
- For each identified flaw in <feedback>, apply the following steps:
    * Identified Flaw: Clearly outline the specific flaw identified in the <feedback> section.
    * Failed Adjustment: Quote the corresponding **Failed Adjustment** of the **Identified Flaw** in <feedback>.
    * Proposed Adjustment: Specify the exact OPERATIONS to address the **Identified Flaw**. Do not reintroduce or reword the same solution in **Failed Adjustment**.
\end{lstlisting}
\end{tcolorbox}

\begin{tcolorbox}[colback=white, colframe=black, title=
Prompt for Learning from the Global Best  
$\mathrm{LLM}_{\text{glob}}$
, width=\linewidth, halign=left, enhanced jigsaw, breakable]
\begin{lstlisting}[language=TeX, basicstyle=\ttfamily\footnotesize, numbers=none, xleftmargin=0pt, frame=none]
You are a strategic assistant tasked with improving a team's performance by analyzing the strengths of a higher-performing example team. Your objective is to understand the specific practices and configurations of the more optimized team that are directly relevant to solving the current team's issues. You will suggest practical improvements to the current team without copying outright.

You are tasked with improving the current team's roles and collaborative workflow:
<current team>
{current_team}
</current team>

This team is designed to solve the following types of tasks:
<task>
{task}
</task>

However, the <current team>'s performance is insufficient and must be improved based on the following feedback:
<feedback>
{feedback}
</feedback>

You have been provided with details of a globally recognized high-performing team, optimized specifically for solving the same type of <task> as the <current team>:
<global best team>
{g_best}
</global best team>

# Instruction

Follow the instructions to generate your response:
- Use the following OPERATIONS to refine roles within the <current team>: 
    * Add Role: Introduce a new role when an existing subtask becomes overly complex or burdensome, requiring a specialized responsibility that cannot be integrated into current roles without disrupting their primary responsibilities. Define the role's:
        - Name: A clear name that reflects its specific responsibility.
        - Responsibility: Specific tasks or functions the role will handle.
        - Policy: Operational guidelines for fulfilling the role's duties.
    * Modify Role: Adjust the policy of an existing role for improved role execution, when the identified inefficiencies or gaps can be addressed through manageable refinements to its policy, ensuring the changes do not overburden the role and are within the scope of its responsibility.
    * Delete Role: Remove roles that are redundant, unnecessary, or conflict with the team's primary objectives.
- Use the following OPERATIONS to optimize the workflow of the <current team>:
    * Add Step: Add a new step if a gap exists in the workflow that hinders overall efficiency, coordination, or goal achievement. Ensure the new step does not duplicate the functions of existing steps and adds clear value to the process. Define the step's:
        - Role: The role responsible for acting in this step.
        - Input: The input for this step must be the output produced by one or more roles in previous steps.
        - Output: What output is expected from the role in this step?
    * Modify Input: Adjust the input of an existing workflow step to ensure that it comprehensively incorporates outputs from previous steps to support the current step.
    * Modify Output: Modify the output of an existing workflow step to ensure that it fully aligns with the expected deliverables of the step and supports the inputs of subsequent steps.
    * Delete Step: Delete a step if it has become redundant, no longer contributes to team goals, or overlaps with other steps in the workflow. Ensure the removal of the step does not impact other steps' efficiency or completeness in achieving objectives.
    * Re-order Steps: Re-order steps if their current sequence causes inefficiencies or coordination issues within the workflow. Ensure the new order improves logical flow without compromising the integrity or dependencies of other steps.
- For each identified flaw in <feedback>, apply the following steps:
    * Identified Flaw: Clearly outline the specific flaw identified in the <feedback> section.
    * Thought: What can we learn from the <global best team>'s descriptions to do better in the **Identified Flaw**?
    * Comparative Insights: 
        - Extract specific elements from the <global best team>'s descriptions that demonstrate excellence in the **Identified Flaw**. 
        - Present these elements as part of a structured sentence, explicitly quoting the key phrases from their role responsibilities, role policies, step inputs, step outputs, or step orders. 
        - Ensure the response integrates the quoted descriptions into a coherent sentence without adding commentary, assumptions, or analysis. 
        - If nothing helpful to solve the **Identified Flaw**, say 'None'.
    * Proposed Adjustment: The adjustment must directly reflect and utilize the specific phrases quoted in the **Comparative Insights**. The wording and content of the adjustment must align with these insights without introducing unrelated suggestions, rephrased ideas, or unquoted elements. The response must clearly demonstrate how the adjustment directly incorporates the practices described in **Comparative Insights**. If **Comparative Insights** is 'None', say 'None' here.
\end{lstlisting}
\end{tcolorbox}

\begin{tcolorbox}[colback=white, colframe=black, title=
Prompt for Learning from the Personal Best  
$\mathrm{LLM}_{\text{pers}}$
, width=\linewidth, halign=left, enhanced jigsaw, breakable]
\begin{lstlisting}[language=TeX, basicstyle=\ttfamily\footnotesize, numbers=none, xleftmargin=0pt, frame=none]
You are a strategic assistant tasked with improving a team's performance by analyzing the strengths of a higher-performing example team. Your objective is to understand the specific practices and configurations of the more optimized team that are directly relevant to solving the current team's issues and to suggest practical improvements to your team without copying outright.

You are tasked with improving the current team's roles and collaborative workflow:
<current team>
{current_team}
</current team>

This team is designed to solve the following types of tasks:
<task>
{task}
</task>

However, the <current team>'s performance is insufficient and must be improved based on the following feedback:
<feedback>
{feedback}
</feedback>

You are provided with the following "personal best team", identified as the most effective setup for addressing the <feedback> throughout the sequence of adjustments made from the initial team setup to the <current team>. This "personal best team" captures the optimal roles and workflow that have proven most successful in solving similar <feedback>, serving as a refined benchmark for guiding improvements to the <current team>'s performance.
<personal best team> 
{p_best} 
</personal best team>

# Instruction

Follow the instructions to generate your response:
- Use the following OPERATIONS to refine roles within the <current team>: 
    * Add Role: Introduce a new role when an existing subtask becomes overly complex or burdensome, requiring a specialized responsibility that cannot be integrated into current roles without disrupting their primary responsibilities. Define the role's:
        - Name: A clear name that reflects its specific responsibility.
        - Responsibility: Specific tasks or functions the role will handle.
        - Policy: Operational guidelines for fulfilling the role's duties.
    * Modify Role: Adjust the policy of an existing role for improved role execution, when the identified inefficiencies or gaps can be addressed through manageable refinements to its policy, ensuring the changes do not overburden the role and are within the scope of its responsibility.
    * Delete Role: Remove roles that are redundant, unnecessary, or conflict with the team's primary objectives.
- Use the following OPERATIONS to optimize the workflow of the <current team>:
    * Add Step: Add a new step if a gap exists in the workflow that hinders overall efficiency, coordination, or goal achievement. Ensure the new step does not duplicate the functions of existing steps and adds clear value to the process. Define the step's:
        - Role: The role responsible for acting in this step.
        - Input: The input for this step must be the output produced by one or more roles in previous steps.
        - Output: What output is expected from the role in this step?
    * Modify Input: Adjust the input of an existing workflow step to ensure that it comprehensively incorporates outputs from previous steps to support the current step.
    * Modify Output: Modify the output of an existing workflow step to ensure that it fully aligns with the expected deliverables of the step and supports the inputs of subsequent steps.
    * Delete Step: Delete a step if it has become redundant, no longer contributes to team goals, or overlaps with other steps in the workflow. Ensure the removal of the step does not impact other steps' efficiency or completeness in achieving objectives.
    * Re-order Steps: Re-order steps if their current sequence causes inefficiencies or coordination issues within the workflow. Ensure the new order improves logical flow without compromising the integrity or dependencies of other steps.
- For each identified flaw in <feedback>, apply the following steps:
    * Identified Flaw: Clearly outline the specific flaw identified in the <feedback> section.
    * Thought: What can we learn from the <personal best team>'s descriptions to do better in the **Identified Flaw**?
    * Comparative Insights: 
        - Extract specific elements from the <personal best team>'s descriptions that demonstrate excellence in the **Identified Flaw**. 
        - Present these elements as part of a structured sentence, explicitly quoting the key phrases from their role responsibilities, role policies, step inputs, step outputs, or step orders. 
        - Ensure the response integrates the quoted descriptions into a coherent sentence without adding commentary, assumptions, or analysis. 
        - If nothing helpful to solve the **Identified Flaw**, say 'None'.
    * Proposed Adjustment: The adjustment must directly reflect and utilize the specific phrases quoted in the **Comparative Insights**. The wording and content of the adjustment must align with these insights without introducing unrelated suggestions, rephrased ideas, or unquoted elements. The response must clearly demonstrate how the adjustment directly incorporates the practices described in **Comparative Insights**. If **Comparative Insights** is 'None', say 'None' here.
\end{lstlisting}
\end{tcolorbox}

\begin{tcolorbox}[colback=white, colframe=black, title=
Prompt for Velocity Update  
$\mathrm{LLM}_{\text{vel}}$
, width=\linewidth, halign=left, enhanced jigsaw, breakable]
\begin{lstlisting}[language=TeX, basicstyle=\ttfamily\footnotesize, numbers=none, xleftmargin=0pt, frame=none]
You are tasked with optimizing a multi-agent team setup to enhance its performance in solving a specific task.

The team to optimize is as follows, including its roles and collaborative workflow:
<current team>
{team}
</current team>

This team is designed to solve the following types of tasks:
<task>
{task}
</task> 

# Objective

Develop a detailed adjustment plan focused on optimizing roles and the collaborative workflow to maximize the <current team>'s performance in addressing the specified <task>. The adjustments must be based on the following feedback:
<feedback> 
{feedback} 
</feedback>

# Instruction

Follow the instructions to generate your response:
- Use the following OPERATIONS to refine roles within the <current team>: 
    * Add Role: Introduce a new role when an existing subtask becomes overly complex or burdensome, requiring a specialized responsibility that cannot be integrated into current roles without disrupting their primary responsibilities. Define the role's:
        - Name: A clear name that reflects its specific responsibility.
        - Responsibility: Specific tasks or functions the role will handle.
        - Policy: Operational guidelines for fulfilling the role's duties.
    * Modify Role: Adjust the policy of an existing role for improved role execution, when the identified inefficiencies or gaps can be addressed through manageable refinements to its policy, ensuring the changes do not overburden the role and are within the scope of its responsibility.
    * Delete Role: Remove roles that are redundant, unnecessary, or conflict with the team's primary objectives.
- Use the following OPERATIONS to optimize the workflow of the <current team>:
    * Add Step: Add a new step if a gap exists in the workflow that hinders overall efficiency, coordination, or goal achievement. Ensure the new step does not duplicate the functions of existing steps and adds clear value to the process. Define the step's:
        - Role: The role responsible for acting in this step.
        - Input: The input for this step must be the output produced by one or more roles in previous steps.
        - Output: What output is expected from the role in this step?
    * Modify Input: Adjust the input of an existing workflow step to ensure that it comprehensively incorporates outputs from previous steps to support the current step.
    * Modify Output: Modify the output of an existing workflow step to ensure that it fully aligns with the expected deliverables of the step and supports the inputs of subsequent steps.
    * Delete Step: Delete a step if it has become redundant, no longer contributes to team goals, or overlaps with other steps in the workflow. Ensure the removal of the step does not impact other steps' efficiency or completeness in achieving objectives.
    * Re-order Steps: Re-order steps if their current sequence causes inefficiencies or coordination issues within the workflow. Ensure the new order improves logical flow without compromising the integrity or dependencies of other steps.
- For each Identified Flaw in <feedback>, apply the following steps:
    * Identified Flaw: Clearly outline the specific Identified Flaw in the <feedback> section.
    * Proposed Adjustment: Based on the **Recommended Adjustment**, **Best Team Insights**, and **Past Best Setup Reflection**, generate a final adjustment plan that directly addresses the **Identified Flaw**, while avoiding any repetition of the **Failed Adjustments**. 
\end{lstlisting}
\end{tcolorbox}

\begin{tcolorbox}[colback=white, colframe=black, title=
Prompt for Position Update  
$\mathrm{LLM}_{\text{pos}}$
, width=\linewidth, halign=left, enhanced jigsaw, breakable]
\begin{lstlisting}[language=TeX, basicstyle=\ttfamily\footnotesize, numbers=none, xleftmargin=0pt, frame=none]
You are an expert assistant and writer. You are tasked with generating a refined team from an existing team according to the reflection.

You are given the roles within the current team:
<roles>
{roles}
</roles>

You are also provided with the workflow of the current team:
<workflow>
{workflow}
</workflow>

The team is solving the following types of tasks:
<task>
{task}
</task>

# Instruction

Your job is to update the roles and workflow of the team based on the following plan:
<plan>
{plan}
</plan>

Use these guidelines when generating the answer:
<system-guidelines>
1. If a role does not require modification in the plan, it must be retained in the final "roles" list with its original "Name," "Responsibility," and "Policy."
2. If the plan specifies that a role should be modified, only update the "Policy"; do not change the "Name" or "Responsibility."
3. If the plan specifies that a role should be removed, then remove it from the final "roles" list.
4. If the plan specifies adding a new role, include it in the final "roles" list with its "Name," "Responsibility," and "Policy."
5. When generating the final answer, verify the total number of roles to ensure:
   - All roles that do not require modification remain unchanged.
   - Roles marked for removal are actually removed.
   - Newly added roles appear in the final list.
   - Modified roles are correctly updated.
6. The information flow must be strictly modular, with each step primarily receiving structured input from the outputs of previous steps. Steps can refer to the initial task definition implicitly as needed, but it should not be treated as a direct input for workflow dependencies.
7. Each step's output must be structured and usable as a direct input for subsequent steps, creating a clear, step-by-step workflow.
8. Each step can only be assigned to a single role and cannot involve multiple roles simultaneously.
9. The final step in the workflow must produce the exact deliverable specified in the <task> without referencing any intermediate steps.
</system-guidelines>
\end{lstlisting}
\end{tcolorbox}

\section{Case Study}
\label{app: Case Study}
\subsection{Illustrative Optimization Process}
\label{app: Optimization Mechanism Illustration}
The following example demonstrates how each component of the velocity update—Global Best, Personal Best, and Failure-Driven Adjustments—contributes to system refinement during the optimization process. The Global Best guidance led to the first major improvement by introducing a dedicated Quality Assurance Specialist role, ensuring structural completeness and consistency across the generated travel plan. The Personal Best guidance triggered the second improvement by identifying a missing verification step between accommodation planning and downstream modules. To address this, a cross-validation step was added, enabling the Accommodation Coordinator to verify constraints such as budget, minimum stays, child suitability, and room availability before forwarding data. Finally, the Failure-Driven Adjustment mechanism refined the Quality Assurance module by incorporating budget compliance checks, directly addressing prior execution failures related to cost violations.

This example illustrates how each optimization signal enables targeted refinements, jointly driving the emergence of well-structured, constraint-compliant agentic workflows. Personal Best Guidance identified a missing verification step between accommodation planning and downstream tasks. Without cross-checking accommodations against user requirements, there was a risk of passing forward incomplete or non-compliant data. To resolve this, a new workflow step was added in which the Accommodation Coordinator performs a cross-verification of accommodation options to confirm alignment with constraints such as budget, minimum stays, child suitability, and room count.

\begin{lstlisting}[
    caption={Example optimization process of the Personal Best Guidance with  \text{LLM}$_{\text{pers}}$},      
    label={lst: codenames},
    basicstyle=\ttfamily\small,
    backgroundcolor=\color{lightgray!20},      breaklines=true,
    breakatwhitespace=true,
    keepspaces=true
]
// previous team
{
    "roles": [
        ...
    ],
    "workflow": [
        ...,
        {
    "Step": "2",
    "Role": "Accommodation Coordinator",
    "Input": "Transportation plan detailing the chosen mode of travel.",
    "Output": "Accommodation plan including the number of nights, recommended hotels, confirmation of minimum stay requirements, suitability for children, and the number of rooms needed."
  },
  {
    "Step": "3",
    "Role": "Restaurant Advisor",
    "Input": "Accommodation plan including the number of nights and recommended hotels, along with cuisine preferences and dietary restrictions from travelers.",
    "Output": "Restaurant recommendations for each non-travel day."
  }
    ]
}
// personal best guidance. Output of LLM_pers
{
    "Identified Flaw": "Missing Workflow Step: There is a missing workflow step between Step 2 and Step 3 that should involve a final review or cross-verification of accommodation options against user requirements before the Quality Assurance Specialist review.",
    "Thought": "Incorporating a cross-verification step can prevent oversights and ensure accommodation options align with requirements before further review.",
    "Comparative Insights": "The personal best team includes a validation step to 'Verify minimum stay requirements for all accommodations,' ensuring a thorough check before final approval.",
    "Proposed Adjustment": "Add a new workflow step after Step 2 where the Accommodation Coordinator performs a cross-verification of accommodation options against user requirements before proceeding to Quality Assurance Specialist."
}

// updated team
{
    "roles": [
        ...,
        {
        "Name": "Accommodation Coordinator",
        "Responsibility": "Plan accommodations for the trip, including the number of nights and suitable hotels.",
        "Policy": "1. Review the travel itinerary and number of nights in each city. 2. Research suitable accommodations based on budget and preferences. 3. Verify minimum stay requirements for all accommodations. 4. Confirm the number of rooms needed for all travelers and ensure accommodations align with user constraints. 5. Provide a primary and a secondary accommodation option for each city, explicitly stating whether the selected accommodations meet the minimum stay requirements and suitability for children."
        }
        ...
    ],
    "workflow": [
        ...
        {
    "Step": "2",
    "Role": "Accommodation Coordinator",
    "Input": "Transportation plan detailing the chosen mode of travel.",
    "Output": "Accommodation plan including the number of nights, recommended hotels, confirmation of minimum stay requirements, suitability for children, and the number of rooms needed."
  },
        {
        "Step": "3",
        "Role": "Accommodation Coordinator",
        "Input": "Accommodation plan including the number of nights, recommended hotels, and user constraints.",
        "Output": "Cross-verified accommodation options confirming adherence to all specified constraints."
        },
        {
    "Step": "4",
    "Role": "Restaurant Advisor",
    "Input": "Accommodation plan including the number of nights and recommended hotels, along with cuisine preferences and dietary restrictions from travelers.",
    "Output": "Restaurant recommendations for each non-travel day."
  }
    ]
}
\end{lstlisting}

In this case, Global Best Guidance identified a gap in the original workflow: the absence of a final review step to ensure all travel constraints and requirements were met. While the Travel Plan Integrator emphasized consistency, it lacked a dedicated validation phase.
To resolve this, a new Quality Assurance Specialist role was added. This role systematically reviews the entire travel plan—covering transportation, accommodations, dining, and attractions—to catch errors and ensure compliance before finalization.
\begin{lstlisting}[
    caption={Example optimization process of the Global Best Guidance with \text{LLM}$_{\text{glob}}$},      
    label={lst: codenames},
    basicstyle=\ttfamily\small,
    backgroundcolor=\color{lightgray!20},      breaklines=true,
    breakatwhitespace=true,
    keepspaces=true
]
// previous team
{
    "roles": [
        [
        {
            "Name": "Transportation Planner",
            "Responsibility": "...",
            "Policy": "..."
        },
        {
            "Name": "Accommodation Coordinator",
            "Responsibility": "...",
            "Policy": "..."
        },
        {
            "Name": "Restaurant Advisor",
            "Responsibility": "...",
            "Policy": "..."
        },
        {
            "Name": "Attraction Specialist",
            "Responsibility": "...",
            "Policy": "..."
        },
        {
            "Name": "Travel Plan Integrator",
            "Responsibility": "...",
            "Policy": "..."
        }
    ]
    ],
    "workflow": [
        ...
    ]
}
// global best guidance. Output of LLM_glob
{
    "Identified Flaw": "The travel plan may lack a final review step to ensure all constraints and requirements are fully met before confirmation.",
    "Thought": "The global best team incorporates a review process to ensure the final travel plan is comprehensive and consistent, suggesting the potential benefit of a dedicated Quality Assurance Specialist role.",
    "Comparative Insights": "The Travel Plan Integrator's policy emphasizes the importance of a review: 'Review the entire plan for consistency and completeness.'",
    "Proposed Adjustment": "Introduce a new Quality Assurance Specialist role responsible for reviewing the entire travel plan to ensure compliance with all constraints and requirements before finalization."
}
//updated team
{
    "roles": [
        ...
  {
    "Name": "Travel Plan Integrator",
    "Responsibility": "...",
    "Policy": "..."
  },
  {
    "Name": "Quality Assurance Specialist",
    "Responsibility": "Review the entire travel plan to ensure compliance with all constraints and requirements before finalization.",
    "Policy": "1. Examine the transportation plan for conflicts and alignment with the itinerary. 2. Verify that accommodations meet all minimum stay and user constraints. 3. Ensure restaurant recommendations align with traveler preferences and dietary restrictions. 4. Confirm attraction selections fit within the planned schedule. 5. Provide feedback for adjustments if necessary."
  }
    ],
    "workflow": [
        ...
    ]
}
\end{lstlisting}

In this case of Failure-Driven Adjustment, the system identified a recurring issue: the Quality Assurance Specialist lacked explicit policy guidelines for verifying the accuracy of outputs from preceding roles. This gap continued to result in budget discrepancies and incomplete itineraries, even after a prior adjustment. The initial revision introduced a general review process, but it lacked a clear, enforceable structure and proved ineffective. Building on this insight, the improved adjustment implemented a mandatory verification process supported by a checklist, ensuring that all outputs are thoroughly reviewed for accuracy and completeness before the finalization of the travel plan. This refinement transforms a vague recommendation into a structured and enforceable policy, significantly enhancing the integrity of the final output.

\begin{lstlisting}[
    caption={Example optimization process of the Failure-Driven Adjustments with \text{LLM}$_{\text{fail}}$},      
    label={lst: codenames},
    basicstyle=\ttfamily\small,
    backgroundcolor=\color{lightgray!20},      breaklines=true,
    breakatwhitespace=true,
    keepspaces=true
]
// previous team
{
    "roles": [
        ...,
        {
    "Name": "Quality Assurance Specialist",
    "Responsibility": "Review the entire travel plan to ensure compliance with all constraints and requirements before finalization.",
    "Policy": "1. Examine the transportation plan for conflicts and alignment with the itinerary. 2. Verify that accommodations meet all minimum stay and user constraints. 3. Ensure restaurant recommendations align with traveler preferences and dietary restrictions. 4. Confirm attraction selections fit within the planned schedule. 5. Provide feedback for adjustments if necessary."
    }
        ...
    ],
    "workflow": [
        ...
    ]
}
//Failure-Driven Adjustment. Output of LLM_fail.
{
"Identified Flaw": "Role Policy Deficiency: The policy for the Quality Assurance Specialist lacks specific guidelines for verifying the accuracy of outputs from previous steps, which could prevent budget discrepancies and incomplete itineraries.",
"Failed Adjustment": "Modify Role: Revise the Quality Assurance Specialist's policy to include a review process for cross-verifying outputs from all roles before finalizing the travel plan. The revised policy will state: '3. Review all outputs from previous roles for consistency and completeness before finalizing the travel plan.'",
"Proposed Adjustment": "Modify Role: Clarify the Quality Assurance Specialist's policy to mandate a verification process that includes a checklist to ensure all outputs from previous roles are accurate and complete before finalization."
}
// updated team
{
    "roles": [
        ...,
        {
    "Name": "Quality Assurance Specialist",
    "Responsibility": "Review the entire travel plan to ensure compliance with all constraints and requirements before finalization.",
    "Policy": "1. Examine the transportation plan for conflicts and alignment with the itinerary. 2. Verify that accommodations meet all minimum stay and user constraints. 3. Ensure restaurant recommendations align with traveler preferences and dietary restrictions. 4. Confirm attraction selections fit within the planned schedule. 5. Provide feedback for adjustments if necessary. 6. Verification of all components, including accommodations, restaurants, and attractions, to confirm alignment with user constraints and comply with budgetary limits."
    }
        ...
    ],
    "workflow": [
        ...
    ]
}
\end{lstlisting}

\section{Best-Discovered Agentic System}
\label{app: Discovered Agentic System}
In this section, we present the final agentic system discovered by SwarmAgentic. These optimized systems—spanning MGSM, Creative Writing, Meeting Scheduling, and TravelPlanner—demonstrate the flexibility and generality of SwarmAgentic in generating task-adaptive agentic structures across diverse domains.

\subsection{MGSM}
\begin{lstlisting}[language=Python]
def forward(team):


    # Step 1: Problem Analysis Specialist analyzes the problem and produces a structured summary of the problem components.
    problem_summary = team.call(
        'Problem Analysis Specialist',
        [],
        "Structured summary of the problem components.",
    )

    # Step 2: Mathematical Operations Specialist uses the structured summary to create a detailed outline of calculations needed to solve the problem.
    calculation_outline = team.call(
        'Mathematical Operations Specialist',
        [problem_summary],
        "Detailed outline of calculations required to solve the problem.",
    )

    # Step 3: Quality Assurance Specialist reviews the detailed outline for accuracy.
    reviewed_outline = team.call(
        'Quality Assurance Specialist',
        [calculation_outline],
        "Reviewed assumptions and interpretations ready for verification.",
    )

    # Step 4: Quality Assurance Specialist verifies the reviewed outline for execution readiness.
    verified_operations = team.call(
        'Quality Assurance Specialist',
        [reviewed_outline],
        "Verified operations ready for execution with corrections if necessary.",
    )

    # Step 5: Calculation Execution Specialist executes the verified operations and returns the final result.
    final_result = team.call(
        'Calculation Execution Specialist',
        [verified_operations],
        "Final result of the calculations.",
    )

    # Step 6: Solution Integration Specialist formats the final result as the final answer.
    formatted_answer = team.call(
        'Solution Integration Specialist',
        [final_result],
        "Formatted final answer.",
    )

    # Return the final formatted answer
    return formatted_answer
\end{lstlisting}

\subsection{Creative Writing}
\begin{lstlisting}[language=Python]
def forward(team):
    # Step 1: Sentence Analyzer analyzes sentences for thematic connections.
    categorized_sentences = team.call('Sentence Analyzer', [], 'Categorized sentences with themes and narrative roles.')
     
    # Step 2: Narrative Architect creates a narrative framework based on the categorized sentences.
     narrative_framework = team.call('Narrative Architect', [categorized_sentences], 'Narrative framework outlining the placement of each sentence.')
    
    # Step 3: Narrative Coherence Reviewer reviews the narrative framework for thematic coherence.
    coherence_feedback = team.call('Narrative Coherence Reviewer', [narrative_framework], 'Feedback on thematic coherence of the narrative framework.')
     
    # Step 4: Feedback Integrator revises the framework based on feedback received.
    revised_narrative_framework = team.call('Feedback Integrator', [coherence_feedback], 'Revised narrative framework ready for paragraph development, detailing how transitions have been integrated.')
    
   # Step 5: Thematic Integration Specialist enhances the thematic integration of the revised framework.
    enhanced_thematic_integration = team.call('Thematic Integration Specialist', [revised_narrative_framework], 'Enhanced thematic integration of the narrative framework.')
    
    # Step 6: Integration Clarity Review confirms readiness for discussion with the Paragraph Developer. 
    clarity_confirmation = team.call('Integration Clarity Review', [enhanced_thematic_integration], 'Confirmation of thematic continuity and readiness for discussion.')
    
    # Step 7: Integrated Feedback Review produces a comprehensive review document.
    review_document = team.call('Integrated Feedback Review', [revised_narrative_framework, clarity_confirmation], 'Comprehensive review document that captures all necessary adjustments.')
     
    # Step 8: Feedback Review Discussion clarifies feedback integration details for paragraph writing.
    discussion_outcome = team.call('Feedback Review Discussion', [review_document], 'Clarified feedback integration details for paragraph writing.')
     
    # Step 9: Paragraph Developer writes the paragraphs based on the integrated feedback.
    final_paragraphs = team.call('Paragraph Developer', [revised_narrative_framework], 'Four concise paragraphs that demonstrate clear thematic coherence and emotional depth.')
    
    # Step 10: Final Integrator reviews the paragraphs and produces the final cohesive narrative document.
    final_narrative = team.call('Final Integrator', [final_paragraphs], 'Final cohesive narrative document, including a comprehensive evaluation of coherence issues.')
    
    # Return the final narrative as the answer.
    return final_narrative
\end{lstlisting}

\subsection{Meeting Scheduling}
\begin{lstlisting}[language=Python]
def forward(team):
    # Step 1: Friend Locator identifies and lists all friends, their locations, and available times.
    friends_list = team.call(
        'Friend Locator',
        [],
        "[{Friend: Name, Location: Place, TimePeriod: [Start, End]}, ...]",
    )

    # Step 2: Travel Time Estimator calculates the travel time between each friend's location.
    travel_times = team.call(
        'Travel Time Estimator',
        [friends_list],
        "[{From: LocationA, To: LocationB, TravelTime: Time}, ...]",
    )

    # Step 3: Travel Time Verifier verifies all travel and waiting times.
    verified_travel_times = team.call(
        'Travel Time Verifier',
        [travel_times],
        "Verified travel and waiting times list",
    )

    # Step 4: Waiting Time Validator reviews validated data for waiting times.
    validated_waiting_times = team.call(
        'Waiting Time Validator',
        [verified_travel_times],
        "Validated list of waiting times.",
    )

    # Step 5: Final Integrator ensures all travel and waiting times are adjusted properly.
    adjusted_times = team.call(
        'Final Integrator',
        [validated_waiting_times],
        (
            "Adjusted travel and waiting times list that resolves discrepancies before scheduling."
        ),
    )

    # Step 6: Meeting Time Optimizer develops a schedule to meet as many friends as possible.
    meeting_schedule = team.call(
        'Meeting Time Optimizer',
        [friends_list, adjusted_times],
        (
            "Finalized meeting schedule that incorporates all validated "
            "travel and waiting times, including a detailed breakdown."
        ),
    )

    # Step 7: Schedule Validator reviews the final meeting schedule for feasibility.
    validated_schedule = team.call(
        'Schedule Validator',
        [meeting_schedule],
        "Validated meeting schedule document.",
    )

    # Return the final validated meeting schedule document.
    return validated_schedule
\end{lstlisting}

\subsection{TravelPlanner}
\begin{lstlisting}[language=Python]
def forward(team):
    # Step 1: Transportation Planner creates a transportation schedule.
    transportation_schedule = team.call(
        "Transportation Planner",
        [],
        (
            "Transportation schedule detailing mode of transport "
            "for each leg of the journey."
        ),
    )

    # Step 2: Accommodation Coordinator creates an accommodation plan based on the transportation schedule.
    accommodation_plan_initial = team.call(
        "Accommodation Coordinator",
        [transportation_schedule],
        "Accommodation plan including number of nights and recommended hotels.",
    )

    # Step 3: Accommodation Coordinator verifies transportation details and user-specific requirements regarding accommodations.
    verified_transportation_details = team.call(
        "Accommodation Coordinator",
        [transportation_schedule],
        (
            "Verified transportation details and user-specific requirements regarding accommodations."
        ),
    )

    # Step 4: Accommodation Coordinator finalizes the accommodation plan, including user preferences and verified details.
    accommodation_plan_final = team.call(
        "Accommodation Coordinator",
        [accommodation_plan_initial],
        "Accommodation plan including user preferences and verified details.",
    )

    # Step 5: Restaurant Advisor recommends restaurants for each non-travel day based on the accommodation plan and user cuisine preferences.
    restaurant_recommendations = team.call(
        "Restaurant Advisor",
        [accommodation_plan_final, verified_transportation_details],
        "Restaurant recommendations for each non-travel day.",
    )

    # Step 6: Attraction Specialist recommends attractions for each day of the trip.
    attraction_recommendations = team.call(
        "Attraction Specialist",
        [accommodation_plan_initial],
        "Attraction recommendations for each day of the trip.",
    )

    # Step 7: Quality Assurance Specialist verifies all components, ensuring constraints are met.
    qa_verification = team.call(
        "Quality Assurance Specialist",
        [
            accommodation_plan_final,
            restaurant_recommendations,
            attraction_recommendations,
        ],
        "Verification of all components ensuring constraints are met.",
    )

    # Step 8: Travel Plan Integrator compiles all components into a comprehensive travel plan.
    comprehensive_travel_plan = team.call(
        "Travel Plan Integrator",
        [
            transportation_schedule,
            accommodation_plan_final,
            restaurant_recommendations,
            attraction_recommendations,
            qa_verification
        ],
        (
            "Comprehensive travel plan including transportation, "
            "accommodation, dining, and attractions."
        ),
    )

    # Return the final comprehensive travel plan.
    return comprehensive_travel_plan
\end{lstlisting}

\subsection{Comparison with ADAS-Discovered Agentic System}
\label{app: Comparison with ADAS-Discovered Agentic Systems}
\begin{figure}
    \centering
    \includegraphics[width=.7\linewidth]{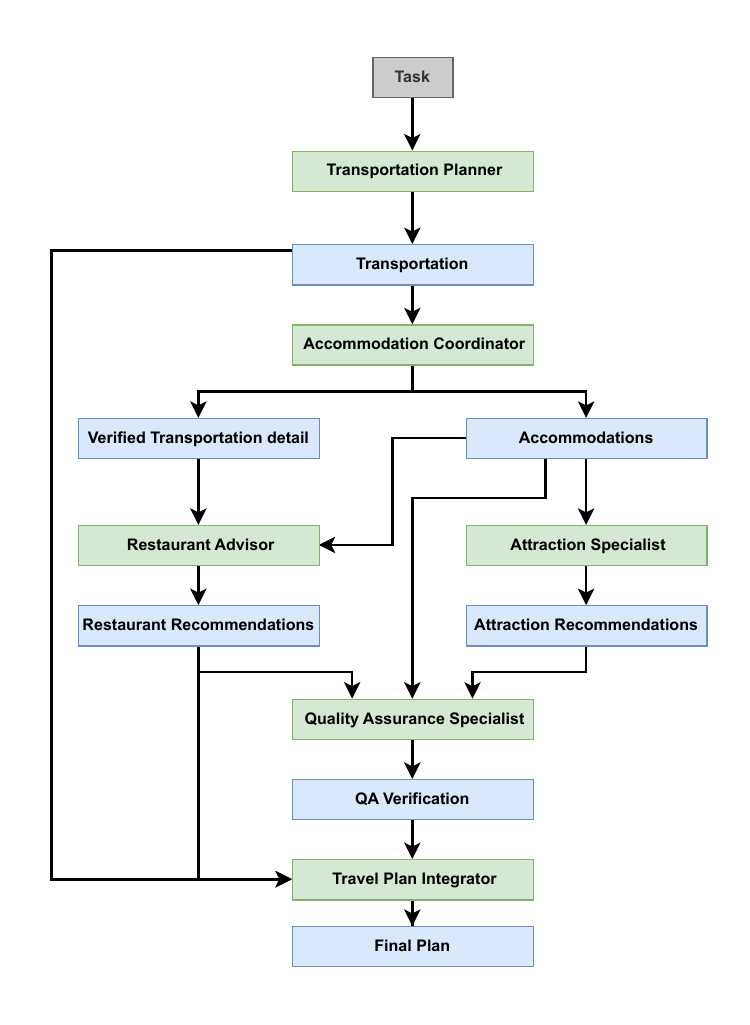}
    \caption{Best agentic system generated by SwarmAgentic for the TravelPlanner task, illustrating the optimized agent roles and coordination structure.}
    \label{fig:traj}
\end{figure}
\begin{figure}
    \centering
    \includegraphics[width=.95\linewidth]{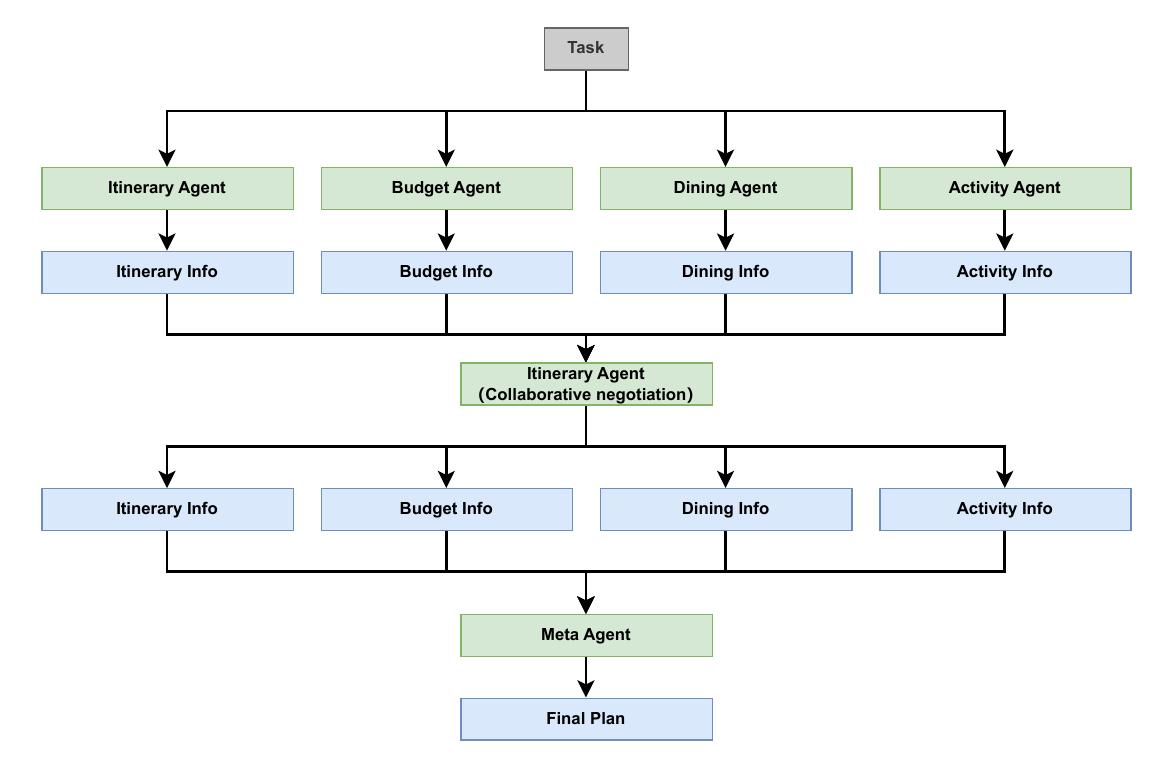}
    \caption{Final agentic system generated by ADAS for the TravelPlanner task, illustrating the optimized agent roles and coordination structure.}
    \label{fig:traj}
\end{figure}

Because Meta Agent Search keeps every previously-generated workflow in its archive, each new prompt handed to the language model must include a long, ever-growing list of full workflow definitions. The sheer size and structural complexity of this archive, as well as the irrelevant details that inevitably accumulate over many search iterations, quickly exhaust the model’s context window and muddle its reasoning. The search algorithm itself prioritizes novelty over optimization, relying on a straightforward strategy aimed at discovering new and potentially interesting designs. Consequently, the search process of ADAS tends to enumerate limitless possibilities within the search space, making it difficult to identify the truly optimal workflow.

As shown in Code 1, the ADAS-discovered workflow assigns distinct roles to specialized agents, including itinerary, budget, dining, activity, and meta agents. However, a critical limitation is the absence of a dedicated accommodation agent. As a result, the system fails to reliably address constraints such as room type, minimum night stays, occupancy limits, and rule-specific conditions (e.g., pet-friendliness). These constraints are central to many travel-related queries, yet no agent is explicitly responsible for enforcing them. Furthermore, the system exhibits difficulty reasoning over multiple constraints jointly, while individual agent proposals may satisfy some conditions, the final plan often violates global requirements due to a lack of coherent integration by the meta-agent.

\begin{lstlisting}[language=Python, caption={Code 1: Optimal workflow generated for TravelPlanner by ADAS}, label={lst:adas}]
def forward(self, taskInfo):
    # Role definitions for specialized agents
    itinerary_instruction = "Please create a detailed travel itinerary considering the constraints and preferences."
    budget_instruction = "Please analyze the budget and suggest accommodations and activities that fit within the budget."
    dining_instruction = "Please suggest restaurants or dining options that match the user's cuisine preferences."
    activity_instruction = "Please recommend activities or attractions based on the user's interests and location."
    dialogue_instruction = "Discuss your proposals with other agents, highlighting strengths and negotiating improvements."

    # Instantiate specialized agents
    itinerary_agent = LLMAgentBase(['thinking', 'itinerary'], 'Itinerary Planner')
    budget_agent = LLMAgentBase(['thinking', 'budget'], 'Budget Manager')
    dining_agent = LLMAgentBase(['thinking', 'dining'], 'Dining Advisor')
    activity_agent = LLMAgentBase(['thinking', 'activity'], 'Activity Coordinator')
    meta_agent = LLMAgentBase(['thinking', 'final_plan'], 'Meta Decision Agent')

    # Gather initial proposals from specialized agents
    itinerary_info = itinerary_agent([taskInfo], itinerary_instruction)[0]
    budget_info = budget_agent([taskInfo], budget_instruction)[0]
    dining_info = dining_agent([taskInfo], dining_instruction)[0]
    activity_info = activity_agent([taskInfo], activity_instruction)[0]

    # Collaborative negotiation phase among agents
    proposals = [itinerary_info, budget_info, dining_info, activity_info]
    for i, proposal in enumerate(proposals):
        for j, other_proposal in enumerate(proposals):
            if i != j:
                dialogue = itinerary_agent([taskInfo, proposal, other_proposal], dialogue_instruction)
                # Update the proposal based on feedback from other agents
                proposals[i] = dialogue[1]  # Assuming the updated proposal comes in the second position

    # Prepare responses for meta-agent
    meta_instruction = "Evaluate the following proposals and create a cohesive final travel plan:"
    final_thinking, final_plan = meta_agent(proposals, meta_instruction)

    return final_plan

\end{lstlisting}

\end{document}